\documentclass{article}
\usepackage{multirow}
\usepackage{changepage}
\usepackage[title]{appendix}
\usepackage[version=3]{mhchem} % Formula subscripts using \ce{}
\usepackage{multirow}
\usepackage{caption}
\usepackage{appendix}
\usepackage{float}
\usepackage[a-1b]{pdfx}

\usepackage{float}
\usepackage{amssymb}
\PassOptionsToPackage{dvipsnames}{xcolor}
\usepackage{xcolor}
\usepackage{xspace}
\usepackage[most]{tcolorbox}
\usepackage{colortbl}
\usepackage{booktabs}
\usepackage[numbers]{natbib}
\usepackage{authblk}
\usepackage{graphicx} % For \resizebox

\usepackage{xr-hyper}
\usepackage{hyperref}

\usepackage[most]{tcolorbox}
\usepackage{colortbl}
\usepackage{booktabs}
\usepackage{amsmath,amssymb}

\usepackage{authblk}

\title{OneProt: Towards Multi-Modal Protein Foundation Models}
\date{}
\author[1]{Klemens Flöge}
\author[2]{Srisruthi Udayakumar}
\author[3,4]{Johanna Sommer}
\author[10]{Marie Piraud}
\author[5,6]{Stefan Kesselheim}
\author[5,7]{Vincent Fortuin}
\author[3,7,4]{Stephan Günnemann}
\author[8,9]{Karel J van der Weg}
\author[8,9]{Holger Gohlke}
\author[10]{Erinc Merdivan}
\author[5,6,$\ast$]{Alina Bazarova}

\affil[1]{PriorLabs, work done at Helmholtz AI, Germany}
\affil[2]{Independent Researcher, Coimbatore, Tamil Nadu, 641014, India}
\affil[3]{School of Computation, Information and Technology, Technical University of Munich, Garching, 85748, Germany}
\affil[4]{Munich Data Science Institute, Technical University of Munich, Garching, 85748, Germany}
\affil[5]{Helmholtz AI, Germany}
\affil[6]{Jülich Supercomputing Centre, Forschungszentrum Jülich, 52425 Jülich, Germany}
\affil[7]{Munich Center for Machine Learning, Munich, 80538, Germany}
\affil[8]{Institute of Bio- and Geosciences (IBG-4: Bioinformatics), Forschungszentrum Jülich, 52425 Jülich, Germany}
\affil[9]{Institute for Pharmaceutical and Medicinal Chemistry, Heinrich Heine University Düsseldorf, 40225 Düsseldorf}
\affil[10]{Helmholtz Munich, 85764 Neuherberg, Germany}
\affil[$\ast$]{Corresponding author: \href{mailto:al.bazarova@fz-juelich.de}{al.bazarova@fz-juelich.de}}

\begin{document}
\maketitle
%%%%%%%%%%%%%%%%%%%%%%%%%%%%%%%%%%%%%%%%%%%%%%%%%%%%%%%%%%%%%%%%%%%%%
%% The "tocentry" environment can be used to create an entry for the
%% graphical table of contents. It is given here as some journals
%% require that it is printed as part of the abstract page. It will
%% be automatically moved as appropriate.
%%%%%%%%%%%%%%%%%%%%%%%%%%%%%%%%%%%%%%%%%%%%%%%%%%%%%%%%%%%%%%%%%%%%%
% \begin{tocentry}

% Some journals require a graphical entry for the Table of Contents.
% This should be laid out ``print ready'' so that the sizing of the
% text is correct.

% Inside the \texttt{tocentry} environment, the font used is Helvetica
% 8\,pt, as required by \emph{Journal of the American Chemical
% Society}.

% The surrounding frame is 9\,cm by 3.5\,cm, which is the maximum
% permitted for  \emph{Journal of the American Chemical Society}
% graphical table of content entries. The box will not resize if the
% content is too big: instead it will overflow the edge of the box.

% This box and the associated title will always be printed on a
% separate page at the end of the document.

% \end{tocentry}

%%%%%%%%%%%%%%%%%%%%%%%%%%%%%%%%%%%%%%%%%%%%%%%%%%%%%%%%%%%%%%%%%%%%%
%% The abstract environment will automatically gobble the contents
%% if an abstract is not used by the target journal.
%%%%%%%%%%%%%%%%%%%%%%%%%%%%%%%%%%%%%%%%%%%%%%%%%%%%%%%%%%%%%%%%%%%%%
\begin{abstract}
Recent advances in Artificial Intelligence have enabled multi-modal systems to model and translate diverse information spaces. Extending beyond text and vision, we introduce OneProt, a multi-modal Deep Learning model for proteins that integrates structural, sequence, text, and binding site data. 
Using the ImageBind framework, OneProt aligns the latent spaces of protein modality encoders in a lightweight fine-tuning scheme that focuses on pairwise alignment with sequence data, rather than requiring full matches. This novel approach comprises a mix of Graph Neural Networks and transformer architectures. It demonstrates  good performance in retrieval tasks and showcases the efficacy of multi-modal systems in Protein Machine Learning through a broad spectrum of downstream baselines, including enzyme function prediction and binding site analysis. Furthermore, OneProt enables the transfer of representational information from specialized encoders to the sequence encoder, enhancing capabilities for distinguishing evolutionarily related and unrelated sequences and exhibiting representational properties where evolutionarily related proteins align in similar directions within the latent space. In addition, we extensively investigate modality ablations to identify the encoders that contribute the most to predictive performance, highlighting the significance of the binding site encoder, which has not been used in similar models previously. This work expands the horizons of multi-modal protein models, paving the way for transformative applications in drug discovery, biocatalytic reaction planning, and protein engineering.
%It demonstrates strong performance in retrieval tasks and surpasses state-of-the-art methods on several biologically relevant downstream tasks, such as enzyme function prediction, binding site analysis, as well as enhanced capabilities of learning distinct representations for evolutionary related and unrelated sequences. This work expands multi-modal capabilities in protein models, paving the way for applications in drug discovery, biocatalytic reaction planning, and protein engineering.
\end{abstract}

%%%%%%%%%%%%%%%%%%%%%%%%%%%%%%%%%%%%%%%%%%%%%%%%%%%%%%%%%%%%%%%%%%%%%
%% Start the main part of the manuscript here.
%%%%%%%%%%%%%%%%%%%%%%%%%%%%%%%%%%%%%%%%%%%%%%%%%%%%%%%%%%%%%%%%%%%%%
\section{Introduction}
The protein space is vast and high-dimensional; even for a 100-residue polypeptide chain, there are $\sim10^{130}$ possible sequences \cite{vila2020protein}. The topological space of functional and synthesizable proteins is significantly smaller, but identifying these meaningful subspaces using experimental methods remains challenging. While machine learning methods for investigating proteins are not new \cite{cheng_machine_2008}, recent increases in computing power \cite{Kuhlman2019Advances}, advances in algorithms, and the availability of extensive sequence data \cite{uniprot_consortium_uniprot_2021} have collectively sparked a revolution in computational protein design \cite{Xu2020, Gligorijevic2021Structure, Verkuil2022}. However, challenges such as achieving tunable control over protein conformations and ensuring precise shape complementarity for molecular recognition are becoming feasible only now \cite{Kortemme2024}.

In recent years, multi-modal Artificial Intelligence (AI) systems and foundation models have notably become more prominent. Initially introduced for text-to-image tasks, the CLIP framework \cite{radford2021learning}, which efficiently learns visual concepts from natural language supervision, has been adapted to various architectures. This evolution is evident in the transition from the multi-modal capabilities of GPT-3 \cite{openai2022intro} to GPT-4 \cite{openai2023gpt}. 
%While the CLIP framework \citet{radford2021learning} focused on aligning models for two modalities, the 
Furthermore, ImageBind \cite{girdhar2023imagebind} demonstrated that aligning pairs of modalities is sufficient to unify the latent space of all modalities, provided one of the paired modalities is consistently present.

While powerful, sequence-only protein language models can struggle to capture functional properties dictated by three-dimensional structure or explicit evolutionary constraints. As highlighted in \cite{Notin2024}, extending models to include additional information beyond sequences is a promising direction for learning richer protein representations.  Integrating complementary modalities addresses these gaps: tools for sequence alignment \cite{steinegger2017mmseqs2} provide evolutionary context, and binding site prediction \cite{p2rank} directly identifies key interaction interfaces. The enriched multi-modal representations support a range of downstream tasks, including protein function prediction \cite{vanderWeg2024} and enzyme or antibody design \cite{Notin2024}, thereby advancing functional protein design \cite{Madani_LLM_functional}.

Building on the success of evolutionary scale modeling (ESM) \cite{esmpaper, MSAtransformer, ESM3}, recent studies show that integrating these models with modality-specific encoders enhances performance in protein-related tasks.  This multi-modal synergy is powerful because it allows a model to, for instance, jointly reason about a residue's sequence context, its structural environment, and its evolutionary conservation. This leads to better performance in tasks such as enzyme function prediction and antibody stability \cite{Zhao_CLEAN,Harmalkar2023}, showcasing the potential of multi-modal approaches for accurate protein function and interaction predictions. In parallel, large-scale sequence-based architectures such as AIDO \cite{AIDOProtein16B} demonstrate the scalability of mixture-of-experts designs, which, while unimodal, reflect the same principle of distributed specialization that underlies multi-modal fusion. However, it is worth noting that such expansive models trained on vast amounts of data have faced criticism, highlighting that an efficient codebase and careful curation of training data can yield better results even with smaller models \cite{amplify}. Moreover, large models are expensive to train and often require significant optimization to perform inference.

%The intersection of foundational biological models and multi-modal AI is a rapidly advancing field of research. Multi-modal training examples encompass a variety of data types, including audio \cite{guzhov2022audioclip}, video \cite{xu2021videoclip}, tabular data \cite{hager2023best}, and point clouds \cite{zhang2022pointclip}. 

Multi-modal AI approaches have been explored in several innovative ways in the molecular sciences, including 3D molecule-text modalities in language models \cite{li2024towards}, the combination of molecule graphs with natural language \cite{su2022molecular}, and text-protein sequence alignment \cite{xu2023protst}. Structural information has been successfully incorporated into sequence vocabularies \cite{Saprot}. In addition, retrieval systems have been developed to query molecular structures based on text descriptions \cite{edwards2021text2mol} and to generate target molecules based on protein pockets \cite{peng2022pocket2mol}. The models developed by \cite{xiao2024molbind}, \cite{protrek} are built using the ImageBind framework. The latter, ProTrek, focuses on all possible pairs of modalities during training and demonstrates retrieval results superior to classical bioinformatics search engines, delivering strong performance in several downstream tasks, using exclusively transformer-based architectures. %We note, however, that ProTrek is using exclusively transformer-based architectures as encoders and ablation studies are not covered by that work. 
On the other hand, \cite{wang2024biobridge} employs a knowledge graph-based approach for modality alignment.

%Successful molecule generation guided by language has been demonstrated using various approaches, such as the work by \citet{liu2023multi} and latent diffusion models \cite{zhu3MDiffusion}. Additionally, molecule-based text generation is exemplified in the works of \citet{liu2023molca} and \citet{li2024towards}. Protein encoders have been utilized in large language models to enhance drug editing capabilities \cite{liu2023chatgptpowered}.

\begin{figure}[h!]%[H]
\center
   \includegraphics[width=0.65\textwidth]{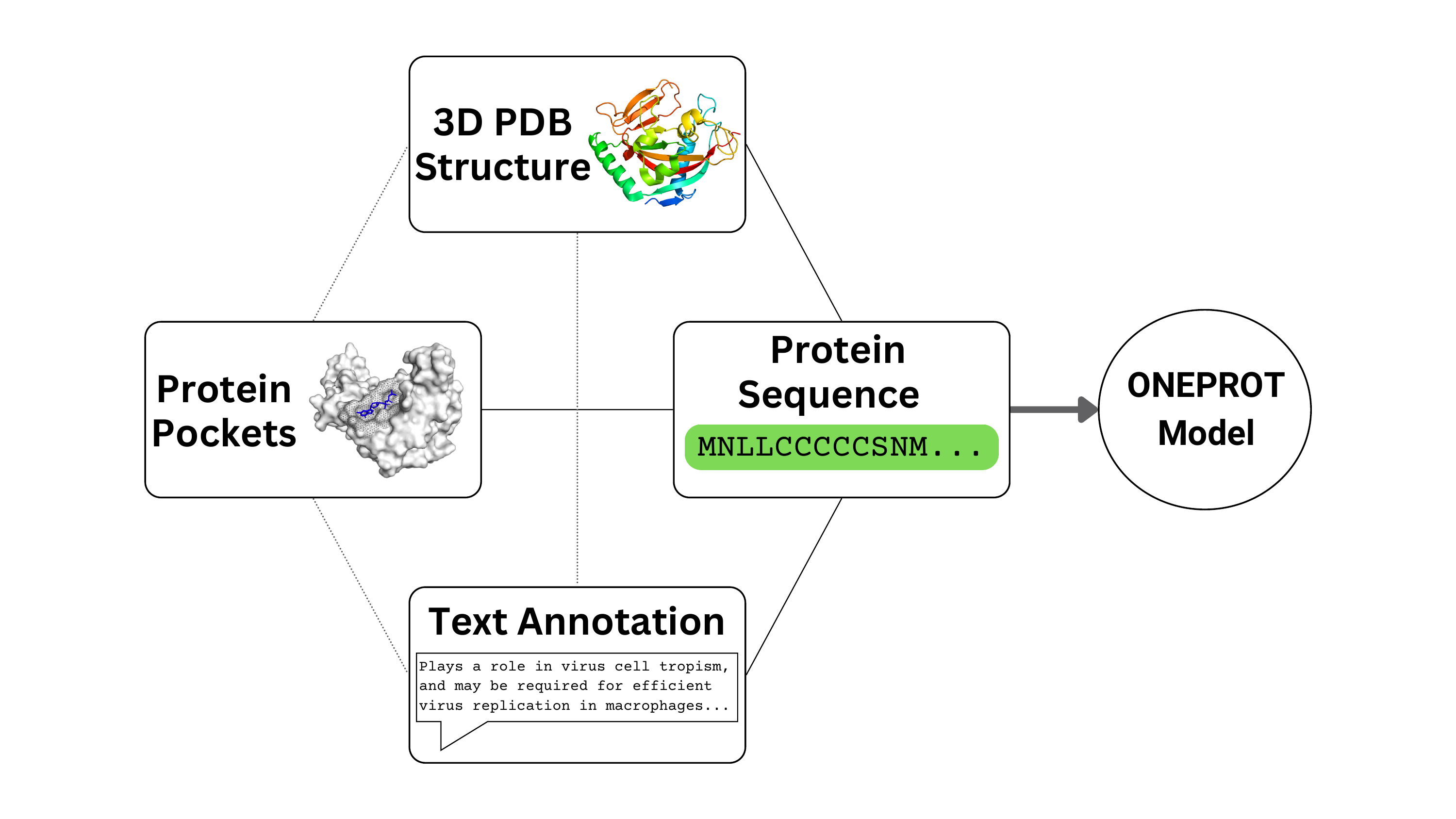}
    \caption{{\bf Overview of OneProt’s alignment of protein sequences with other modalities for comprehensive cross-modal integration.}%function descriptions, structures, and pockets 
    Training is performed using pairs comprising a sequence and another modality, leading to the emergent alignment between all other modalities, as indicated by the dashed lines.
    }
    \label{fig: imagebind illustration}
\end{figure}

In this manuscript, we introduce OneProt, Fig~\ref{fig: imagebind illustration}, which effectively extends the ImageBind framework to the protein space using transformer-based and Graph Neural Network (GNN) encoders. Our model aligns protein sequences, protein structures represented in two ways, protein pockets, and text annotations by pairing each of the latter modalities with the sequence modality only. We curate the respective training datasets using publicly available databases. Moreover, to reduce redundancy, we focus on aligning the core modalities by clustering the sequences at $\leq50\%$ identity. OneProt's multi-modal alignment enables efficient, unbiased retrieval and downstream tasks using a comprehensive protein dataset. Its flexible and versatile framework also allows one to easily add new modalities to the model, if necessary. We conduct extensive ablations to evaluate each modality's contribution to downstream tasks, addressing a gap left by similar models. Based on performance, we introduce two models: OneProt-5 (five modalities) and OneProt-4 (four modalities). Importantly, their moderate size allows for a seamless application to new downstream tasks when needed.

\section{Materials and methods}
\label{oneprotmodel}

\subsection*{Representation alignment}\label{subsec: representation alignment}
\begin{figure*}[!h]
    \centering
 \includegraphics[width= 0.8\textwidth]{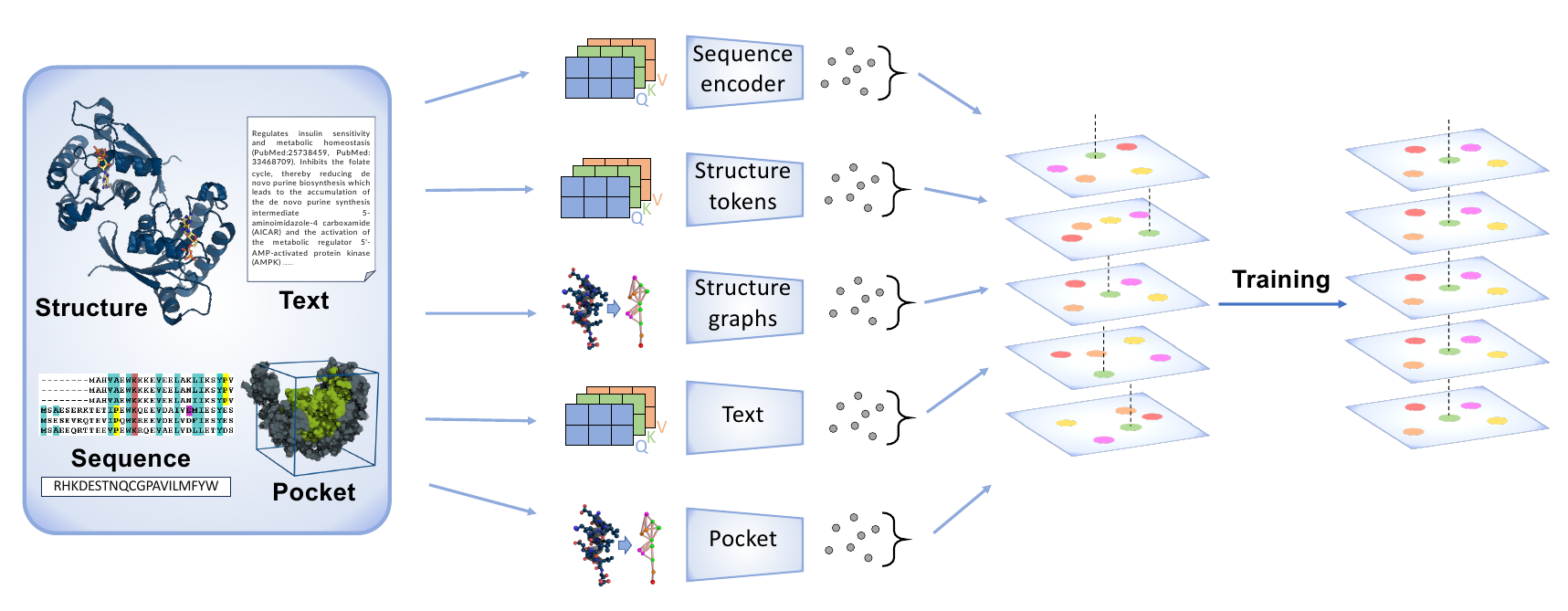}
    \caption{{\bf Overview of the OneProt model.} The model aligns multiple modalities, including primary protein sequence, 3D protein structure, binding pockets, and text annotations. Each modality is processed by its respective encoder, generating embeddings aligned in a shared latent space, facilitating cross-modal learning and integration.}
    \label{fig: oneprot overview}
\end{figure*}

Illustrated in Figs~\ref{fig: imagebind illustration} and \ref{fig: oneprot overview}, the training data of the OneProt model consists of paired samples from multiple modalities, each contributing unique information for a comprehensive protein representation. In modality pairs $(\mathcal{F}, \mathcal{E})$, $\mathcal{F}$ always denotes the protein sequence, which is consistently present in each tuple, while $\mathcal{E}$ can represent the structure, text, or pocket modality. A sample pair $(a_i, b_i)$ consists of data points from two different modalities of the same protein, where $i\in\{1\dots n\}$ are the indices within a batch of size $n$. Representations $\mathbf{a'}_i$ and $\mathbf{b'}_i$ are obtained using their respective encoders $\phi_{\mathcal{F}}$ and $\phi_{\mathcal{E}}$: $\mathbf{a'}_i = \phi_{\mathcal{F}}(a_i)$ and $\mathbf{b'}_i = \phi_{\mathcal{E}}(b_i)$, where $\phi_\mathcal{F}: \mathcal{F} \rightarrow \mathbb{R}^m$, and $\phi_\mathcal{E}:\mathcal{E} \rightarrow \mathbb{R}^k$, and therefore $\mathbf{a'}_i$ and $\mathbf{b'}_i$ are numeric vectors of the dimensions $m$ and $k$, $m$ not necessarily equal $k$. Then, in order to align the latent spaces of the encoders, projection heads $proj_{\mathcal{F}}: \mathbb{R}^m\rightarrow \mathbb{R}^l$ and $proj_{\mathcal{E}}: \mathbb{R}^k\rightarrow \mathbb{R}^l$ are applied to $\mathbf{a'}_i$ and $\mathbf{b'}_i$ respectively, thereby mapping them int the shared space $\mathbb{R}^l$. The projected vectors $proj_{\mathcal{F}}(\mathbf{a'_i})$ and  $proj_{\mathcal{E}}(\mathbf{b'_i})$ are then $L_2$-normalized to produce final unit embeddings $\mathbf{a}_i=proj_{\mathcal{F}}(\mathbf{a'_i})/||proj_{\mathcal{F}}(\mathbf{a'_i})||_2$ and  $\mathbf{b}_i=proj_{\mathcal{E}}(\mathbf{b'_i})/||proj_{\mathcal{E}}(\mathbf{b'_i})||_2$. 
%The encoded samples $\mathbf{a}_i$ and $\mathbf{b}_i$ reside in the corresponding latent spaces.

%To align these latent spaces of the encoders $\phi_\mathcal{F}: \mathcal{F} \rightarrow \mathbb{R}^l$ and $\phi_\mathcal{E}:\mathcal{E} \rightarrow \mathbb{R}^l$, the latent space dimension $l$ must be the same. 
Given a batch of pairs $\{(a_1, b_1), ..., (a_n, b_n)\}$, the goal is to synchronize the representations of $\mathbf{a}_i$ and $\mathbf{b}_i$ (positive pairs) while pushing $\mathbf{a}_i$ and $\mathbf{b}_j$ apart for $i \ne j$ (negative pairs). 
%Following \citet{radford2021learning}, the InfoNCE loss \cite{infonce} is used for this with $L_1$ regularization:
Following \cite{radford2021learning}, the InfoNCE loss \cite{infonce} is defined as:

\setlength{\abovedisplayskip}{9pt}
\setlength{\belowdisplayskip}{9pt}

\begin{equation}
L_{\mathcal{F}, \mathcal{E}}=-\frac1n\sum_i\log \frac{\exp \left(\mathbf{a}_i^{\top} \mathbf{b}_i / \tau\right)}{\exp \left(\mathbf{a}_i^{\top} \mathbf{b}_i / \tau\right)+\sum_{j \neq i} \exp \left(\mathbf{a}_i^{\top} \mathbf{b}_j / \tau\right)}%+\frac{\lambda}{l} (||\mathbf{a}_i||_1+||\mathbf{b}_i||_1),
\label{eq: infonce loss} 
\end{equation}

\noindent  where $\mathbf{a}_i^{\top} \mathbf{b}_i$ is a dot-product in $\mathbb{R}^l$ between the normalized vectors $\mathbf{a}_i$ and $\mathbf{b}_i$, reflecting therefore cosine similarity between them, and $\tau$ %and $\lambda$ are 
 is the temperature parameter.  Small values of $\tau$ lead to larger-magnitude arguments in the exponent (logits). As a result, even small differences in similarity between the positive pair and the negatives translate into large differences in probability, enforcing stronger separation. Conversely, large values of $\tau$ shrink the logits, flattening the softmax distribution. This reduces sensitivity to differences in similarity, making the separation between positive and negative pairs less pronounced. The total loss, based on the loss in (\ref{eq: infonce loss}), is computed while considering the order of the modalities as follows:
 %and regularization parameters, respectively. Here, $\mathbf{b}_i$ serves as the in-batch negative sample as described in ref. \cite{radford2021learning}. The total loss is calculated as the symmetric sum of the losses for the two modalities $(\mathcal{F}, \mathcal{E})$:
\setlength{\abovedisplayskip}{5pt}
\setlength{\belowdisplayskip}{5pt}
% Write one sentence for the next equation?!
\begin{equation}
L_{\text{total}} = L_{\mathcal{F}, \mathcal{E}} + L_{\mathcal{E}, \mathcal{F}}.
\label{eq: symmetrical loss}    
\end{equation}
This symmetric formulation ensures bidirectional alignment, so the model learns both to retrieve $\mathcal{E}$ given $\mathcal{F}$ and vice versa, following the CLIP-style contrastive setup.

%Following the ImageBind framework, we iterate through sequential batches of paired modalities $(\mathcal{F}, \mathcal{E}_1), \ldots, (\mathcal{F}, \mathcal{E}_n)$, consistently including the sequence modality $\mathcal{F}$ in each batch  as an anchor, making a gradient step after computing loss for each pair $(\mathcal{F}, \mathcal{E}_i)$. 
In line with the ImageBind framework, we  construct batches of paired modalities $(\mathcal{F}, \mathcal{E}_1), \ldots, (\mathcal{F}, \mathcal{E}_n)$, consistently including the sequence modality $\mathcal{F}$ in each batch as an anchor. For each pair $(\mathcal{F}, \mathcal{E}_i)$ we compute the contrastive loss and immediately apply a gradient update of the respective encoders and projection heads.  This sequential optimization differs from the original CLIP training, which aggregates losses across a fixed two-modality pair (image–text) and updates once per batch. This approach ensures the alignment of the sequence modality with every other modality. Moreover, because all modalities are mapped into the same shared latent space via $\mathcal{F}$, modalities, which were not directly paired together during training, still align with each other - this is also known as emergent alignment, as illustrated in Fig~\ref{fig: oneprot overview}.

\subsection*{Model details}

OneProt integrates pre-trained models across various modalities when available. Otherwise, the corresponding encoder is trained from scratch using the loss as in Eqs (\ref{eq: infonce loss}), (\ref{eq: symmetrical loss}). In what follows, we elaborate on the choice of the respective encoders, their suitability for our tasks, and architectural details. The training methodology is summarized in Table \ref{tab: encoders}: we employ a mix of frozen, fully trained, and Low-Rank Adaptation (LoRA) \cite{lora} methods to optimize performance across modalities. 

\begin{table}[!htbp]
\centering
\caption{Overview of OneProt's different encoders}
\resizebox{1.0\linewidth}{!}{%
\begin{tabular}{@{}rrrrrrr@{}}\toprule
Modality & Model & Training &Pooler &Projection & Full Parameter Count  &Trainable Parameter Count \\ \midrule
Sequence & ESM2 & Freeze & attention & linear   &$652$ M &$0$ \\
Structure-PDB & ProNet & Full & - & linear &  $2.6$ M & $2.6$ M\\
Structure-Token & ESM2 & Full &mean  & linear  &$35$ M &$35$ M\\
%MSA & ESM-1b & Freeze & $117$ M \\
Pocket & ProNet & Full & - & linear & $2.6$ M & $2.6$ M\\ 
Text & MSR BiomedBERT & LoRA & cls &MLP & $110$ M & $0.4$ M\\ \midrule 
Total: &  & & & & $802.2$ M & $40.6$ M \\ 
%Total: &   & $819.4$ M \\  
\end{tabular}%
}
\label{tab: encoders}
\end{table}

The models described in this manuscript and the \nameref{sec:supp} were pre-trained using  64 NVIDIA A100 GPUs (16 compute nodes) of the JUWELS Booster supercomputer \cite{booster} for 33,000 optimizer steps  using Distributed Data Parallel (DDP) scheme. Further information on the pre-training is available in section \ref{s-supp:pretraining}. 

\subsubsection*{Sequence and structure token encoder} 
ESM2 is a transformer-based architecture, which serves as a standard foundation model for the sequence modality, having demonstrated superior performance across a wide range of protein-related tasks \cite{sequencemodel, esmpaper,esm2_strong}. Therefore, we use ESM2 650M \cite{sequencemodel} for the common modality sequence with its pre-trained representations as a stable reference point to anchor the other modalities. 

The respective encoder works as follows. Given a protein sequence \( x \in \mathbb{R}^{n \times d} \), where \( n \) is the sequence length and \( d \) is the embedding dimension, the transformer-based encoder processes the input sequence through layers of self-attention and feed-forward neural networks.

The embedding for the input sequence is initialized as:
\[
x_0 = E(x) + P(x)
\]
where \( E(x) \) denotes the learned embedding for each amino acid in the sequence, and \( P(x) \) represents the positional encoding to capture the sequential nature of the data. This embedding is then processed through multiple layers of the transformer model.

Each transformer layer consists of two main components: multi-head self-attention and a position-wise feed-forward network. For a single transformer layer, the update is given as:
\[
\text{MultiHead}(\{Q_i,K_i,V_i\}_{i=1}^p) = \text{Concat}(\text{head}_1, \text{head}_2, \dots, \text{head}_p) W^O
\]
where \( Q_i = x_0W_i^Q\), \( K_i=x_0W_i^K \), and \( V_i=x_0W_i^V \) represent the query, key, and value matrices, and each head performs scaled dot-product attention:
\[
\text{head}_i = \text{softmax}\left(\frac{Q_iK_i^T}{\sqrt{d^i_k}}\right)V_i
\]
where all $W$ are learnable parameter matrices, and $W^K_i\in \mathbb{R}^{d\times d^i_k}$ . The attention mechanism captures dependencies across the sequence by allowing each position to attend to all others. Following the attention block, the position-wise feed-forward network updates the representation as:
\[
\text{FFN}(h) = \max(0, hW_1 + b_1)W_2 + b_2
\]
where \( W_1 \) and \( W_2 \) are learned weight matrices and $h$ is the output of the multi-head self-attention.

The transformer updates the sequence representation through stacked layers of multi-head self-attention and position-wise feed-forward networks:
\[
h_{\text{seq}}^{(l+1)} = \text{TransformerLayer}(h_{\text{seq}}^{(l)}), \quad l = 1, 2, \dots, L
\]
where \( L \) is the number of layers, and \( h_{\text{seq}}^{(l)} \in \mathbb{R}^{n \times d} \) represents the sequence embeddings at the \( l \)-th layer.

The final output of the sequence transformer is:
\[
h_{\text{seq}} = h_{\text{seq}}^{(L)}
\]
which provides the learned sequence representations for downstream tasks, capturing local and global dependencies across the protein sequence.

To capture the broader structural organization of a protein from a discrete 3D token representation, we used structure tokens as proposed in \cite{Saprot} to encode the structure modality, and trained the aforementioned transformer architecture from scratch, given the lack of a suitable pre-trained model for the structure modality.

%To learn the broader structural organization from a discrete 3D token representation, we used the structure tokens as proposed in \cite{Saprot} to encode the structure modality, and trained the described transformer architecture from scratch in the absense of  a directly applicable pre-trained model, in order to capture structural information in a sequence-like format. 

The Multiple Sequence Alignment (MSA) modality has also been efficiently implemented via the transformer-based ESM-1b model \cite{MSAtransformer}, however, this modality was not included in the final training due to resulting in a significant speed decline and high memory consumption. 

\subsubsection*{Graph structure and pocket encoder}

To allow the network to capture detailed chemical information of the protein structure, we train the all-atom ProNet graph model \cite{wang2023learning}, a  strong performer in graph-based protein modeling, from scratch, due to the absence of a suitable pre-trained model.

The ProNet encoder operates by modeling protein structures as hierarchical 3D graphs to capture relationships at multiple levels of granularity: amino acid, backbone, and all-atom levels. Each amino acid is represented as a node, and edges between nodes are defined by a cut-off radius. This hierarchical representation is particularly valuable for multi-modal learning, as it enables the model to align structural features at different scales with complementary modalities—for instance, potentially matching local atomic interactions with sequence motifs.

A protein graph \( G = (V, E, P) \) is constructed, where \( V \) represents nodes for aminoacids, \( E \) denotes edges for interactions between them, and \( P \) contains the positional information of the atoms in each amino acid. ProNet incorporates a complete geometric representation \( F(G) \), for each level, to effectively learn the hierarchical structures of proteins. At the amino acid level, ProNet constructs a coarse-grained representation using the coordinates of  \( C_{\alpha} \) atoms, capturing the amino acid's overall position and orientation relative to its neighbors.

To represent the all-atom structure, ProNet computes Euler angles (\( \tau_1, \tau_2, \tau_3 \)) between the planes of neighboring amino acids. These angles capture the rotational degrees of freedom between adjacent amino acids. At the all-atom level, side-chain torsion angles (\( \chi_1, \chi_2, \chi_3, \chi_4 \)) provide fine-grained details of each amino acid's side chain.

The hierarchical message passing in ProNet is governed by the equation:
\[
v_i^{l+1} = \text{UPDATE} \left(v_i^l, \sum_{j \in N_i} \text{MESSAGE}(v_j^l, e_{ij}, F(G)) \right)
\]
where \( v_i^l \) is the feature vector of node \( i \) at layer \( l \), \( N_i \) denotes the neighbors of node \( i \), and the \text{UPDATE} and \text{MESSAGE} functions process the node and edge features as well as geometric information from \( F(G) \).

By incorporating the complete representations at multiple levels, ProNet captures global and local structural details of proteins, enabling it to outperform other methods in various protein-related tasks. We use the ProNet implementation as presented in the DIG library \cite{DIG_lib}. To learn the local chemical environment within a protein pocket, we trained an all-atom ProNet model also for the pocket modality from scratch. 

\subsubsection*{Text encoder}

For the text modality, we selected the transformer-based MSR BiomedBERT \cite{Gu_2021}. This model was specifically pre-trained on a large corpus of biomedical literature, making it well-suited to our dataset. It has demonstrated state-of-the-art performance on relevant tasks such as biomedical text classification \cite{ZHOU2025104825}. We employ it to encode protein descriptions for the critical task of mapping them to the standardized controlled vocabulary from UniProt \cite{uniprot_consortium_uniprot_2021}.

%For the text modality, we selected transformer-based MSR BiomedBERT \cite{Gu_2021} because it was specifically trained on biological text descriptions, making it well-suited to our dataset. It is also instrumental in interpreting the controlled vocabulary from UniProt \cite{uniprot_consortium_uniprot_2021}.

The encoder leverages the Masked Language Model (MLM) approach, where input text sequences $x = (x_1, x_2, \ldots, x_n)$ are partially masked, and the model predicts the masked tokens $x_m$. The training objective is to minimize the cross-entropy loss between the predicted tokens $\hat{x}_m$ and the original masked tokens $x_m$:

\[
\mathcal{L}_{\text{MLM}} = - \sum_{m \in \mathcal{M}} \log P(x_m | x_{\setminus m}; \theta),
\]

\noindent where $\mathcal{M}$ denotes the set of masked positions, $x_{\setminus m}$ represents the input sequence with the $m$-th token masked out, and $\theta$ are the model parameters. This technique enables the model to learn deep contextualized representations, $\mathbf{h}_i$, for each token $x_i$. By incorporating domain-specific corpora, such as PubMed abstracts, and using specialized vocabulary derived from in-domain text, the model is fine-tuned to understand complex biomedical language effectively. 

We adapt the text modality to align more closely with the sequence embeddings by applying LoRA, which provides efficient fine-tuning while preserving most of the pre-trained weights. Namely, the weight matrix $W'$ of the network is represented as $W'=W+\frac{\alpha}{r} \Delta W=W+\frac{\alpha}{r} AB$, where $W\in \mathbb{R}^{n\times m}$ is the frozen original weight matrix of the pre-trained model, $A\in\mathbb{R}^{n\times r}$, $B\in\mathbb{R}^{r\times m}$ are the matrices of low rank $r\ll \min(n,m)$, which are updated during the gradient descent algorithm, and $\alpha$ is a scaling parameter, controlling the contribution of the low-rank updates.

%This diverse ensemble of models allows \OneProt to effectively process and integrate information across multiple protein-related modalities.

%In the training methodology, we adopt a multi-modal strategy, carefully calibrating the training approach across different encoders. 
% The ProNet models for structure and pocket modalities, along with the ESM2 model for structure tokens, are fully trained due to the lack of directly applicable pre-trained models. 
\subsubsection*{Projection head}
To enable contrastive learning on the embedding spaces of the same dimension, we implement a fully trainable projection layer on top of each protein encoder's latent space. This ensures that the latent spaces of different encoders are aligned and compatible for downstream tasks. The encoders vary in pooling and projection methods to suit the specifics of each modality, as summarized in Table \ref{tab: encoders}. 

Specifically, for sequence modality, we adopt the Attention1D Pooling Head mechanism, inspired by \cite{tan2024ses-adapter}.  The projection head employs a 1D convolution operation, which can be described as follows: given an input sequence $x \in \mathbb{R}^{N \times L \times C_{\text{in}}}$, where $N$ is the batch size, $L$ is the sequence length, and $C_{\text{in}}$ is the number of input channels, we apply a convolutional filter $w \in \mathbb{R}^{C_{\text{out}} \times C_{\text{in}} \times K}$, where $C_{\text{out}}$ is the number of output channels and $K$ is the kernel size.

The convolution operation for each position $t$ in the sequence and output channel $j$ can be expressed as:

\begin{equation}
y_{t,j} = \sum_{i=1}^{C_{\text{in}}} \sum_{k=0}^{K-1} w_{j,i,k} \cdot x_{t+k,i}
\end{equation}

\noindent where $w_{j,i,k}$ represents the filter weights and $x_{t+k,i}$ is the input at the corresponding position and channel.

To handle variable-length sequences and focus on relevant parts of the input, we incorporate a masked convolution approach. This involves applying a binary mask $m \in \mathbb{R}^{N \times L \times 1}$ to the input:

\begin{equation}
y_{t,j} = \sum_{i=1}^{C_{\text{in}}} \sum_{k=0}^{K-1} w_{j,i,k} \cdot (x_{t+k,i} \cdot m_{t+k})
\end{equation}

This masking technique ensures that the convolution operation only considers valid positions in the sequence, effectively handling padding and improving the model's ability to focus on meaningful information.

The Attention1D Pooling Head further refines the output by applying an attention mechanism, allowing the model to dynamically weight different parts of the sequence based on their importance for the downstream task. 

%All modalities incorporate fully trainable projection layers, standardizing the latent space dimensions across encoders to achieve coherent, unified representations suitable for contrastive learning and downstream applications. 

\subsection*{Data preparation}\label{subsec: data_prep}
Our dataset combines the OpenFold training database \cite{ahdritz2023openproteinset} with UniProtKB/Swiss-Prot \cite{Boutet2007}. Using MMseqs2 \cite{steinegger2017mmseqs2}, we cluster sequences at 50\% identity, to balance, on the one hand, that each cluster represents a homologous group in the protein fold space and, on the other hand, to ensure that clustered proteins share evolutionary relationships. We align the training, validation, and test splits by these sequence clusters. 
%[XXX: This is a very generous seq. ident. cutoff. Why not use 30 proc. or even lower? As there is a structure for each seq., one could have split according to FoldSeek clustering? Is it possible to test this before submission as I fear that a reviewer might ask for it? => I still have my concerns, but let's see ...] 
Leveraging the unique UniProtAC identifier of each protein sequence, we filter the MSAs from OpenFold according to the UniProtACs, and locate a structure in the AlphaFold2DB for each one \cite{alphafolddb}. To obtain the structure tokens, we use the SaProt \cite{Saprot} training dataset, removing sequences with less than $50\%$ sequence identity to the validation and test sets. We then randomly sub-sample a million entries and mask the sequence tokens to extract structure tokens for each UniProtAC. Using P2Rank \cite{p2rank}, we predict the binding site for each structure, where possible, resulting in fewer entries than for the structure modality. We create text annotations by mining the UniProt dataset for the keywords associated with the corresponding UniProtACs. Dataset sizes are listed in Table \ref{tab:dataset}. Full details on dataset creation can be found in the \nameref{sec:supp}, section \ref{appendix_datasets}.

\begin{table}[ht]
\centering
\caption{OneProt Data Overview}
\label{tab:dataset}
\resizebox{0.35\linewidth}{!}{
\begin{tabular}{@{}rr@{}}
\toprule
Modality & Dataset Size \\ \midrule
Sequence & $1.04$ M \\
Structure PDB & $656$ K \\
Structure Tokens & $1$ M \\
Pocket & $341$ K \\ 
Text & $546$ K \\ \bottomrule %\midrule
%TopEnzyme & $231$ K \\ %\bottomrule
\end{tabular}
}
\end{table}

\subsection*{Downstream evaluation}\label{sec:meth_down}

\subsubsection*{Modality alignment}

After aligning modalities (Section \nameref{subsec: representation alignment}) and training OneProt using the symmetrical loss function (Eq (\ref{eq: symmetrical loss})), we evaluate the alignment of latent spaces by constructing a vector database from paired test datasets and conducting cross-modality similarity searches, thereby verifying that representations of the same protein are consistently proximate across different modalities. %Full pre-training details can be found in the supplementary material \ref{s-appendix_datasets}.  
%Instead of the computationally intensive $\mathcal{O}(n^2)$ cosine similarity approach, we leverage Meta's \texttt{FAISS} library \cite{johnson2017billionscale, douze2024faiss} for efficient searches. 

To assess modality alignment, we evaluate each modality's retrieval performance against the sequence reference modality.
For $n$ modalities, we define $2 \times (n-1)$ cross-modal retrieval tasks from the sequence modality, encompassing both forward and reverse directions. Additionally, we introduce emergent retrieval tasks to evaluate modality pairs not directly trained together, resulting in $(n-1)(n-2)$ tasks for $n$ modalities.

 For each modality pair, we compute R@1, R@10, and R@100, where these metrics correspond to the accuracy of the correct retrievals from the nearest, closest ten, or closest hundred embeddings in the latent space, respectively, with the accuracy value of 1 corresponding to the perfect alignment between modalities. We also compute median rank (MR), which measures how well matching representations align by ranking the cosine similarities between pairs modalities for each protein, taking the median of these ranks and averaging across all proteins. Lower values indicate better performance, with 1 representing the best possible score. 
%defined as the median position at which the correct matching item appears when ranking candidates across modalities. 
More details on how MR was computed is available in section \ref{s-supp:pretraining}.

We use 4000 held-out modality pairs for each of the cross-modal retrieval tasks as the test set.\\[5pt]

\subsubsection*{Supervised fine-tuning for downstream tasks}\label{sec:methods:downstream}
We demonstrate the utility of OneProt by evaluating its performance on a range of protein-related downstream tasks, following the ones presented in \cite{Saprot} and using the same metrics, where the accuracy thresholds for classification tasks are fixed and chosen in a standard way. These tasks use well-established benchmarks to encompass protein structure, property, and function predictions. Details on these evaluations can be found in the section \ref{experimentdetails: supervised finetuning}.

For the thermostability task, we use the “Human-cell” splits from FLIP \cite{Dallago2021FLIP}, which were designed to predict protein thermostability values directly from sequence.  This is a regression task, where the evaluation metric is a Spearman's rank correlation coefficient $\rho$ defined for paired observations $(X,Y)=\{(X_i,Y_i)\}_{i=1}^n$, and their respective ranks $(R(X),R(Y))=\{(R(X_i),R(Y_i))\}_{i=1}^n$, with sample size $n$ as follows
\begin{equation}\label{eq:rho}
\rho=\frac{\textrm{cov}(R(X), R(Y))}{\sigma_{R(X)}\sigma_{R(Y)}}
\end{equation}
 The numerator of Eq (\ref{eq:rho}) corresponds to the covariance between the ranks of $(X,Y)$, and the denominator corresponds to the product of the respective standard deviations. The data for this task is drawn from the experimental results reported in the Meltome Atlas \cite{meltome}. %Thermostability, often represented by the melting temperature (Tm), is a crucial property influencing protein function under different environmental conditions, making this task highly relevant in both biological and industrial contexts. 
 
Second, we assess OneProt on the Human Protein-Protein Interaction (HumanPPI) task \cite{Xu2022PEER}. This is a binary classification task for protein pairs: positive pairs are defined as experimentally validated interactions from the Human Protein Reference Database \cite{hprd}, while negative pairs are constructed from proteins localized to different subcellular compartments.
 %Predicting interactions between human proteins is crucial for elucidating cellular networks and pathways, which are vital for understanding disease mechanisms and identifying potential therapeutic targets. 
 
Third,  we evaluate the model's performance on the Metal Ion Binding task \cite{Hu2022Exploring}, which examines a protein's ability to bind metal ions.  This, again, is a binary classification task, where proteins annotated with metal-ion binding sites in the Protein Data Bank \cite{burley2023rcsb} are treated as experimentally confirmed positive instances, and those lacking such annotations serve as negative instances.
%Metal ions are fundamental in stabilizing protein structures and are involved in a range of biological processes, including enzyme catalysis and signal transduction. 

Fourth, we evaluate the prediction of Enzyme Commission (EC) numbers \cite{Gligorijevic2021Structure}. The hierarchical EC classification system enables the prediction of enzyme functions, essential for understanding the catalytic roles of enzymes in biochemical reactions. Fifth, for a key task of functional genomics, Gene Ontology (GO) annotation \cite{Gligorijevic2021Structure}, we evaluate OneProt's ability to predict three types of protein function: molecular function (MF), biological process (BP), and cellular component (CC).  Each of these four tasks is a multilabel classification problem with 585, 489, 1943, and 320 labels, respectively. For the evaluation, we use the $F_{\text{max}}$ score, defined as follows:

\begin{align}\label{eq:fmax}
Precision(\tau)=\frac1N\sum_i^Nprec_i(\tau)\qquad Recall(\tau)=\frac1N\sum_i^Nrec_i(\tau)\nonumber\\
prec_i(\tau)=\frac{TP_i(\tau)}{TP_i(\tau)+FP_i(\tau)}\qquad
rec_i(\tau)=\frac{TP_i(\tau)}{TP_i(\tau)+FN_i(\tau)}\nonumber\\
F_{1}(\tau)=\frac{2Precision(\tau)Recall(\tau)}{Precision(\tau)+Recall(\tau)}\nonumber\\
F_{max}=\max_{\tau}F_1(\tau)
\end{align}
 
In Eq (\ref{eq:fmax}), $i\in\{1\dots N\}$ enumerates proteins in the sample; $TP_i, FP_i$, and $FN_i$ correspond to the number of true positive, true negative, and false negative labels for protein $i$, respectively; $\tau$ is a decision threshold, such that the labels with predicted scores above $\tau$ are classified as positive. %This classification into functional categories is critical for understanding the roles proteins play in complex biological systems.
 
Finally, the DeepLoc benchmark \cite{AlmagroArmenteros2017} was employed to predict the subcellular localization of a protein.  It comprises two tasks: a binary classification task to determine whether a protein is membrane-bound or soluble, and a ten-class classification task that assigns each protein to a specific subcellular location. All proteins used in this downstream evaluation are experimentally annotated in the UniProt database. 
 
For both binary and multiclass classification tasks, we report accuracy as the primary evaluation metric, following the original study \cite{Saprot}. For binary tasks, we additionally report the Area Under the Receiver Operating Characteristic Curve (AUC) to provide complementary evaluation.

As one of the baselines, we include the results of SaProt-LoRa from the original study \cite{Saprot}, under the assumption that they were already optimal; consequently, mean and standard deviation are not reported for this model. This robust baseline fine-tunes a structure-aware sequence model using a Multi-Layer Perceptron (MLP) projection head for supervised learning in each downstream task, thereby updating  7-12 million parameters during the fine-tuning phase, out of which 5.4 million parameters remain in the SaProt backbone. In contrast, our approach is simpler: we keep our baseline ESM/OpenFold/SaProt/ProTrek models and OneProt frozen, generate sequence embeddings, and apply an MLP for the supervised learning problem using these embeddings.  This procedure involves fine-tuning up to 1.5 million parameters, depending on the hyperparameter configuration and the output dimension, which results in faster convergence and lower memory requirements. While our setup is versatile and would support any multi-class classifier, we restrict our analysis to MLPs. This approach only requires downloading the pre-trained OneProt model to generate embeddings, enabling easier application across alternative datasets with reduced computational overhead.
 
  Apart from ESM-2, which is a baseline encoder for our model, we have investigated the performance of a more contemporary and larger model, ESM-3, as well as the ESM-IF \cite{esmif} encoder, which takes protein structures as inputs and returns the respective embeddings. Another baseline is the OpenFold model in the SoloSeq mode \cite{openfold}, which substitutes traditional MSA input by ESM-1b embeddings \cite{esmpaper}. We include OpenFold as a baseline because its structure-focused architecture, offering a complementary comparison to sequence-based encoders. Note, that we provide two tri-modal ProTrek baselines which use the corresponding ESM-2 encoders with 35M and 650M parameters,  overall amounting to 200 and 930 million trainable parameters, respectively, against 40.6 million trainable parameters in the OneProt backbone. Both of these utilize the  contrastive learning framework but were trained on a dataset of 40M datapoints, whereas OneProt's largest modality amounts to 1.04M datapoints. While ProTrek-35M was trained from encoder checkpoints for a duration comparable to OneProt, ProTrek-650M required nearly twice as much training time, with both ProTrek models using a more complex parallelization scheme. We do not compare OneProt’s training time with models other than ProTrek, as these were either trained from scratch, typically requiring orders of magnitude more computational resources, or, as in the case of OpenFold SoloSeq, lack publicly available details on how the ESM-1b checkpoint was adapted.

We evaluate model performance on all downstream tasks over six independent runs with random initializations and report the values corresponding to means and standard deviations across these runs. To statistically compare model performances, we conduct two types of two-sample Wilcoxon rank-sum tests, as the data did not satisfy the normality assumption. Firstly, to assess whether OneProt models outperform the baselines, we apply a one-sided test with the alternative hypothesis that OneProt performs better than the baseline. In this setting, rejecting the null hypothesis at $p < 0.05$ indicates that OneProt significantly outperforms the baseline. Secondly, to evaluate whether OneProt’s performance is comparable to the baselines, we use a two-sided test with the alternative hypothesis that the performance distributions differ. Here, failure to reject the null hypothesis implies that the performance difference between OneProt and the baseline is not statistically significant at the 0.05 level. Hereafter, we report the $p$-values from the one-sided Wilcoxon test when stating that one model outperforms another, and from the two-sided Wilcoxon test when stating that the models have comparable performance.\\[5pt]
 
\subsubsection*{Enzyme function prediction}
To further substantiate our claim that OneProt not only aligns the latent spaces of encoders across various protein data types but also holds potential as a building block for a foundational protein model, we evaluate it on a large downstream task specifically designed to aid enzyme function prediction using the large-scale TopEnzyme database \cite{vanderweg_TopEnzyme}. TopEnzyme is a collection of experimentally resolved  (around 20,000 structures from Binding MOAD database \cite{moad}) and computationally predicted protein structures ordered by EC numbers, consisting of around $231$K enzymes, with a 30\% sequence identity split. EC numbers comprise a four-number hierarchy, where the first three levels represent the main-, sub-, and subsub-class functions, while the fourth level is the specific enzyme function designation. %The dataset is also included in Table \ref{tab:dataset}. 

Similarly to section \nameref{subsec: saprot tasks}, we train a simple MLP based on the embeddings  produced by the baseline and OneProt models to learn and predict EC numbers at the full hierarchy, i.e., to predict the specific enzyme function designation, overall comprising 826 classes. We compare the performance of OneProt models against several strong baselines on the EC task: ESM-2, ESM-IF, OpenFold, and ProTrek-650M, identified as top-performing models in Section~\nameref{subsec: saprot tasks}. We also include ProTrek-35M in our benchmark, as it shares a similar architecture with OneProt and was trained for a comparable duration. As in previous sections, models were evaluated over six independent runs with random initializations, and the predictions from these runs were concatenated to account for variability. We also consider two deep learning models specifically designed to classify EC numbers,
TopEC, \cite{vanderWeg2024}, a graph neural network encoding a local descriptor of protein structure, and CLEAN \cite{Zhao_CLEAN}, a contrastive learning model for sequence data.  For these, we did not perform six independent runs; instead, we report the results provided in the original papers, where the models had already been optimized.

%The results for different classifiers are shown in Figure \ref{fig:topenzyme_boxplot} and are compared to two deep learning models specifically designed to classify EC numbers, TopEC \cite{vanderWeg2024}, a graph neural network encoding a local descriptor of protein structure, and CLEAN \cite{Zhao_CLEAN}, a contrastive learning model for sequence data. 

To assess each model's performance across classes, we compute the Area Under Precision-Recall curve (AUPR) for each class, calculating precision and recall within the instances of that class, and then visualize the distribution  over 6 runs via the corresponding boxplots.

As in the previous section, we use a one-sided two-sample Wilcoxon rank-sum test to evaluate whether OneProt outperforms the baselines (alternative hypothesis) with respect to the AUPR value distributions.\\[5pt]

% Results and Discussion can be combined.
\subsubsection*{Representation learning for evolutionary related proteins}\label{sec:methods:representation_learning}

 Because MSA enables the detection of sequence similarity that correlates with evolutionary relatedness, we leverage the OpenProteinSet \cite{ahdritz2023openproteinset}, which provides extensive MSA data obtained from an all-against-all search on Uniclust30 \cite{steinegger_uniclust}, to evaluate whether OneProt can provide distinct representations of evolutionary relationships compared to the baseline ESM2 and ProTrek models.

%allowing us to evaluate \OneProt's ability to generate biologically meaningful embeddings.

For this, following \cite{hammingmsa}, we process MSA files by computing pairwise Hamming distances to rank sequences based on their similarity or dissimilarity relative to a reference sequence. Analogously to the approach used in \cite{beyond_esm2}, to verify the protein representation, for each reference, we select 50 most similar sequences and 50 most divergent ones, alongside 1,000 unrelated sequences as a control group. Using the ESM-2, ProTrek-35M and -650M, and two OneProt models, we generate embeddings and compute the cosine similarity between each reference sequence and its aligned (similar/divergent) or unrelated sequences. In this representation learning experiment across different models, we refrain from using SaProt, ESM-IF and OpenFold to avoid the inclusion of structural information in the input or architecture, and, therefore, only focus on the protein sequence.

 This task illustrates zero-shot learning, where the models are applied to the dataset without any prior fine-tuning. To compare cosine similarity distributions, we use a one-sided Wilcoxon rank-sum test: for paired samples when comparing embeddings from different models within the same sequence class, and for unpaired samples when comparing embeddings across different sequence classes.\\[5pt]

\subsubsection*{ProSPECCTs}

The ProSPECCTs (Protein Site Pairs for the Evaluation of Cavity Comparison Tools) benchmarking initiative \cite{prospeccts} represents a significant advancement in the field of computational biology, specifically in the analysis and comparison of protein-ligand binding sites. This study meticulously developed a series of tailored benchmark datasets designed to evaluate the performance of various binding site comparison methodologies. The ProSPECCTs datasets encompass a diverse range of protein cavity pairs,  enabling systematic evaluation of the strengths and limitations of different comparison tools across multiple application domains. This comprehensive framework is crucial for guiding researchers in selecting appropriate tools for specific challenges, such as drug re-purposing, promiscuity prediction, or protein-protein interactions analysis. The dataset details are provided in the  Table \ref{s-tab:prospecct_dataset}.  We note, that these datasets comprise both computationally derived and experimentally determined proteins. The latter include crystallographic data (DS1, DS5 in Table \ref{s-tab:prospecct_dataset}), Nuclear Magnetic Resonance spectroscopy data (DS2), and others.  %, with classification difficulty rising with the dataset number. %[XXX: Can one say that the level of difficulty increases with increasing DS number?]

%To assess the performance of OneProt across sequence, structure, and pocket as well as joint modalities, 
Similar to Section \nameref{sec:methods:representation_learning}, we demonstrate the zero-shot capabilities of OneProt here. We evaluate OneProt alongside the top-performing ProTrek models from Section \nameref{sec:methods:downstream}, as well as the ESM-2 model, which serves as the baseline for both ProTrek and OneProt, to highlight the incremental value of incorporating additional modalities. %To assess the performance of sequence embeddings of OneProt models and to compare them to the ones of ESM-2 and ProTrek, we generate the corresponding embeddings for the ProSPECCTs datasets.
After generating ProSPECCTs embeddings from the corresponding models, we compute cosine similarities to group the distinct pairs as defined in the ProSPECCTs dataset. We then apply threshold values to these similarity scores to generate Receiver Operating Characteristic curves, which enable classification of pairs as similar or dissimilar. Finally, we calculate Area Under Receiver Operating Curve (AUC) values.\\[5pt]

\subsubsection*{Ablations}

For all downstream tasks described in the previous sections, we conduct exhaustive ablations, resulting in the evaluation of 15 different models, ESM-2 corresponding to the 16th, sequence-only ablation. We followed the same protocol of 6 independent runs, as in the sections \nameref{subsec: saprot tasks} and \nameref{subsec:topenzyme}.   Moreover, we visualize the performance differences between models using heatmaps for normalized performance drop $\Delta_{x,y}$ between a pair of models $x$ and $y$, which is defined as
\begin{equation}\label{eq:norm_drops}
\Delta_{x,y}=\frac{Perf_y-Perf_x}{Perf_x},
\end{equation}
where $Perf$ corresponds to the corresponding original averaged metric (accuracy in case of binary classification) from the section \nameref{subsec: saprot tasks}. We perform statistical testing as described in section \nameref{sec:methods:downstream}.

\section{Results}\label{experiments}

Here, we discuss the results obtained by performing downstream tasks for two OneProt models, OneProt-5 and OneProt-4, with the structure token encoder omitted in the latter. Moreover, we present the ablations and draw conclusions on the links between the  modality alignment and downstream performance.

\subsection*{Modality alignment}
\label{subsec: modality alignment}

 % representing twice the total number of edges in a fully connected graph with $(n-1)$ nodes (excluding the sequence modality).

\begin{figure*}[!h]
\centering
    \includegraphics[width=\textwidth]{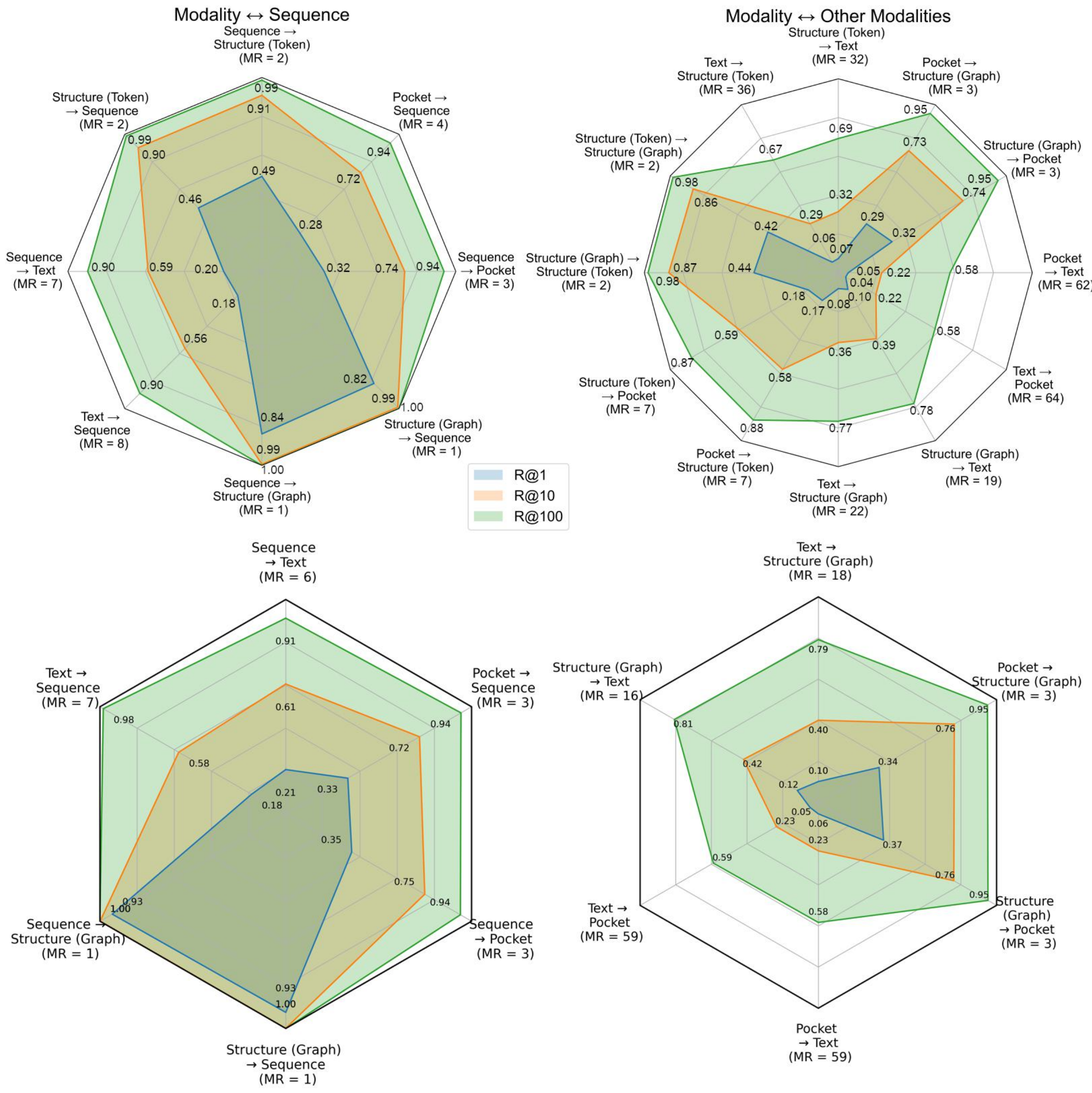}
    \caption{{\bf Alignment performance across modality combinations paired (left column) and not paired (emergent, right column) during training for OneProt-5 (top row) and OneProt-4 (bottom row).} The axes of the polygons correspond to the modality pairs, and the vertices correspond to R@1 (inner polygon), R@10 (middle polygon), and R@100 (outer polygon), which represent the fraction of queries for which the correct (ground-truth) match appears among the top 1, top 10, or top 100 retrieved embeddings, respectively, with the best possible value being equal to 1. MR is the Median Rank of the corresponding embedding in the other modality,  best possible being equal 1.
   % {\bf Alt Text:} Two radar charts illustrating the alignment between different modality pairs. The left one corresponds to modality-sequence, and the right one corresponds to the emergent modalities alignment.
    %[XXX: What does R@100 tell wrt the large number of data points? Could one evaluate, how "similar" the other 99 embeddings in latent space are?] a) The performance from a modality to sequence or vice versa. [XXX: Ok? Vice versa in italics?] b) The emergent performance between unpaired modalities.
    }
    \label{fig:spider_tables}
\end{figure*}

Fig~\ref{fig:spider_tables} (left column) presents the retrieval results for modality pairs trained together for OneProt-5 (top row) and OneProt-4 (bottom row), respectively, showing strong alignment with high R@1 values, especially in sequences paired with structural tokens or graphs. This suggests that the model effectively captures relationships when modalities are explicitly linked during training. Interestingly, while the retrieval performance is good for training modalities, such as sequence paired with structure or graphs, R@1 values remain relatively low for certain modalities, indicating that even with CLIP-style training, perfect synchronization of the encoders is not achieved. This raises the question of whether complete alignment can be attained through pairwise training alone. Furthermore, text alignment performs the worst among the trained modalities, likely due to redundancies in the text data and the inherent complexity of this modality. Fig~\ref{fig:spider_tables} (right column) shows the results for emergent alignments across unpaired modalities.  Note, that OneProt has not seen any of these proteins during training. Generally, the alignment performance is weaker for untrained pairs than trained pairs. Yet, in the worst case, a median rank of 64 is obtained for the task $Text \to Pocket$, which is still considerably lower than the unaligned median rank of 2000 (given the test set size of 4000), providing evidence of an emergent alignment across all tasks. 

We also note, that the corresponding alignment values of the OneProt-4 model are higher than those of OneProt-5, suggesting that more training is required to achieve better results for the model with a higher number of modalities. This is supported by the ablation studies shown in Table \ref{s-tab:ablations}  and can be explained by the growing complexity of the embedding space as the number of modalities increases. Moreover, contrastive loss heavily relies on effective negative sampling. With more modalities, the number of potential negative samples increases significantly, making it harder to distinguish between positive and negative pairs. Future training strategies should aim to enhance the emergent alignments to improve cross-modal adaptability.

\subsection*{Supervised fine-tuning for downstream tasks}
\label{subsec: saprot tasks}

 %Accurate prediction of the subcellular compartment in which a protein resides provides valuable insights into its functional role and its involvement in various cellular processes.

%[XXX: Are the datasets "balanced" if the accuracy is evaluated?]
%[XXX: Can we sort the columns in Table 3 according to the order of the tasks mentioned above? To me, this is an order of increasing "biol. difficulty" of the tasks.]
%[XXX: To what extent could the (generous) sequence identity cut-off of 50\% for developing OneProt "leak through" into the downstream tasks?]

Tables \ref{tab: saprot tasks} and \ref{tab: saprot tasks 1} present the downstream performance of various models across this diverse set of downstream tasks. The evaluation of OneProt-4 and -5 demonstrates its broad applicability in different biological contexts, largely outperforming SaProt  with a frozen backbone, also presented in \cite{Saprot},  with $p<0.01$.

\begin{table*}[!h]
    %\small % Make font smaller
    \renewcommand{\arraystretch}{1.0} 
    
    % Push caption and table to the left by removing centering and using flushleft
    \begin{adjustwidth}{-1.5in}{0in}
    \caption{Performance comparison of OneProt, SaProt-LoRa, SaProt, ProTrek, ESM, and OpenFold on five diverse downstream biological protein tasks: ThermoStability (regression), HumanPPI, Metal Ion Binding, DeepLoc Binary (binary classification) and DeepLoc Subcellular (multiclass classification), using Spearman correlation for ThermoStability, accuracy (ACC) and Area Under the Reciever Operating Curve (AUC) for the remaining tasks.}
    \label{tab: saprot tasks}
    
    % Resize table to fit page width - just slightly smaller
    \resizebox{1.3\textwidth}{!}{%
    \begin{tabular}{lc>{\hspace{0.8cm}}c>{\hspace{0.8cm}}c>{\hspace{0.8cm}}c>{\hspace{0.8cm}}c}
        \hline
        \multirow{3}{*}{\textbf{Model}} & \multirow{2}{*}{\textbf{Thermostability}} & \multirow{2}{*}{\textbf{HumanPPI}} & \multirow{2}{*}{\textbf{Metal Ion Binding}}  & \multicolumn{2}{c}{\textbf{DeepLoc}} \\  \cline{5-6} 
                    &  &    &            & Subcellular & Binary \\
                       & Spearman's $\rho$ & ACC/AUC\%    & ACC/AUC\%           & ACC\% & ACC/AUC\% \\ 
        \hline
\textbf{SaProt-LoRa}    & 0.724         & 86.4       & 75.8               & 85.6       & 93.6       \\
\textbf{SaProt}       & 0.702 (0.005) & 87.1/92.9 (1.4/2.9) & 71.3/76.8 (1.3/1.7)  & 79.0 (0.4) & 91.0/95.6 (0.3/1.4) \\
\textbf{ESM-2}        & 0.696 (0.005) & 86.0/94.2 (1.4/0.6) & 67.4/76.2 (1.3/1.7)  & 81.0 (0.4) & 91.4/96.3 (0.4/0.2) \\
\textbf{ESM-3}        & 0.691 (0.015) & 83.6/92.1 (1.8/1.0) & 72.7/81.8 (1.9/0.9)  & 76.3 (0.5) & 90.8/95.7 (0.2/0.1) \\
\textbf{ESM-IF}       & 0.645 (0.006) & 78.7/86.0 (1.5/0.6) & 69.0/77.7 (1.4/0.6)  & 61.5 (0.6) & 84.7/90.5 (0.5/0.2) \\
\textbf{OpenFold}     & 0.582 (0.013) & 84.4/92.2 (2.3/1.0) & 71.7/76.8 (0.7/1.1)  & 80.0 (0.4) & 91.7/96.6 (0.3/0.2) \\
\textbf{ProTrek-35M}  & 0.638 (0.010)  & 86.4/94.5 (1.5/0.7) & 76.0/94.5 (1.0/0.7) & 83.7 (0.3) & 93.1/97.8 (0.4/0.1) \\
\textbf{ProTrek-650M} & 0.646 (0.009) & 90.2/97.0 (1.5/0.5) & 75.4/82.3 (3.2/0.4)  & 90.9 (0.3) & 95.3/97.0 (0.2/0.5) \\
%\rowcolor{yellow}
\textbf{OneProt-5}    & 0.673 (0.010) & 85.9/93.4 (0.2/0.7) & 76.2/82.1 (1.6/1.3) & 80.3 (0.2) & 92.4/96.3 (0.2/0.2) \\
%\rowcolor{yellow}
\textbf{OneProt-4}    & 0.668 (0.006) & 88.8/95.3 (1.7/0.3) & 77.3/85.3 (0.5/1.0)  & 81.7 (0.4) & 92.1/96.5 (0.3/0.2)\\
        \hline
    \end{tabular}%
    }
    \end{adjustwidth}
\end{table*}

\begin{table*}[!h]
    %\small % Make font smaller
    \renewcommand{\arraystretch}{1.0} 
    
    % Push caption and table to the left by removing centering and using flushleft
    %\begin{adjustwidth}{-2.0in}{0in}
    \caption{Performance comparison of OneProt, SaProt-LoRa, SaProt, ProTrek, ESM, and OpenFold on four multi-label function prediction tasks: Enzyme Commission numbers (EC), Gene Ontology (GO) terms corresponding to Molecular Function (MF), Biological Process (BP), and Cellular Component (CC), using maximum F1-score metric (Fmax) defined by Eq (\ref{eq:fmax})}
    \label{tab: saprot tasks 1}
    
    % Resize table to fit page width - just slightly smaller
    \resizebox{0.9\textwidth}{!}{%
    \begin{tabular}{lc>{\hspace{0.8cm}}c>{\hspace{0.8cm}}c>{\hspace{0.8cm}}c}
        \hline
        \multirow{3}{*}{\textbf{Model}}  & \multirow{2}{*}{\textbf{EC}} & \multicolumn{3}{c}{\textbf{GO}}  \\  \cline{3-5} 
                            &  & \textbf{MF}   & \textbf{BP}   & \textbf{CC}    \\
                            &Fmax  &Fmax &Fmax &Fmax\\
        \hline
\textbf{SaProt-LoRa}          & 0.884         & 0.678         & 0.356         & 0.414                \\
\textbf{SaProt}        & 0.863 (0.004) & 0.623 (0.007) & 0.472 (0.004) & 0.549 (0.004)  \\
\textbf{ESM-2}         & 0.878 (0.003) & 0.645 (0.003) & 0.479 (0.003) & 0.547 (0.004)  \\
\textbf{ESM-3}         & 0.871 (0.004) & 0.643 (0.004) & 0.482 (0.003) & 0.531 (0.010)  \\
\textbf{ESM-IF}        & 0.896 (0.006) & 0.611 (0.005) & 0.437 (0.004) & 0.488 (0.006)  \\
\textbf{OpenFold}      & 0.888 (0.004) & 0.655 (0.005) & 0.491 (0.002) & 0.548 (0.004)  \\
\textbf{ProTrek-35M}   & 0.846 (0.003) & 0.651 (0.002) & 0.514 (0.005) & 0.583 (0.008)  \\
\textbf{ProTrek-650M}  & 0.876 (0.005) & 0.675 (0.006) & 0.538 (0.004) & 0.617 (0.005)  \\
%\rowcolor{yellow}
\textbf{OneProt-5}     & 0.875 (0.005) & 0.656 (0.002) & 0.492 (0.003) & 0.556 (0.005)  \\
%\rowcolor{yellow}
\textbf{OneProt-4}     & 0.871 (0.003) & 0.656 (0.001) & 0.495 (0.003) & 0.555 (0.006) \\
        \hline
    \end{tabular}%
    }
    %\end{adjustwidth}
\end{table*}

As the results in Table \ref{tab: saprot tasks} demonstrate, OneProt models achieve competitive performance across all tasks, regardless of the biological differences inherent to each task. Notably, OneProt attains strong results without relying on the more complex  fine-tuning in the sequence and structure token modalities or augmenting sequence data with structural tokens, as in Saprot, \cite{Saprot}. Moreover, OneProt outperforms the ESM ($p<0.02$ for OneProt-5 and $p<0.006$ for OneProt-4), OpenFold ($p<0.046$ for OneProt-4) and SaProt ($p<0.01$) baselines in most tasks, indicating that contrastive learning across different protein encoders effectively aligns them and transfers representational knowledge across modalities. It also delivers comparable ($p>0.41$ for OneProt-5 on Metal Ion Binding and HumanPPI tasks, two-sided test) to superior ($p<0.006$ for OneProt-4 on HumanPPI, ThermoStability, Metal Ion Binding, EC and GO-MF tasks; $p<0.006$ for OneProt-5 on EC, ThermoStability and GO-MF tasks, one-sided test) results to ProTrek-35M on most tasks, and performs similarly ($p>0.059$ for OneProt-4 on HumanPPI, EC, Metal Ion Binding, and $p>0.57$ for OneProt-5 on EC and Metal Ion Binding, two-sided test) or better ($p<0.002$ on ThermoStability) than ProTrek-650M on a number of tasks. We note, however, that in the case of the Metal Ion Binding task, the average performance of ProTrek-650M is lower than that of OneProt models, while the standard deviation is several times higher. Moreover, in terms of AUC, OneProt-4 outperforms ProTrek-650M with $p<10^{-3}$. Other than that, AUC results mainly align with the accuracy scores, but exhibit narrower distributions, indicating greater robustness. Comprehensive $p$-value results from the two-sample Wilcoxon rank-sum test, including both the one-sided test and the two-sided test, are provided in Figures \ref{s-fig:heatmaps_p}-\ref{s-fig:heatmaps_ablations}. Moreover, for most downstream tasks, the OneProt-4 and OneProt-5 models yield samples with a lower interquartile range (IQR) compared to the baseline models, as summarized in Tables \ref{s-fig:iqr1}-\ref{s-fig:iqr4}, indicating robustness of the OneProt results. 

The OneProt results are achieved using a substantially smaller pre-training dataset, where modalities are paired only with the sequence modality, more closely reflecting real-world scenarios in which datasets across modalities often differ in size or have limited overlap. This underscores OneProt's data efficiency and demonstrates, that considerably longer training, as in case of ProTrek-650M (discussed in \nameref{oneprotmodel}) does not always lead to significant performance gains. We attribute this to the fact that the additional, graph-based structure and pocket modalities of OneProt provide the information required to achieve good performance on the downstream tasks. Of note, is also that the model OneProt-4, which contains only the GNN encoder for the structure, delivers outstanding performance on the HumanPPI, Metal Ion Binding, GO-MF and -BP tasks and competitive performance across other tasks. Although it is still slightly inferior in performance to OneProt-5 on a number of the other tasks, this suggests that the GNN structure encoder may in general be sufficient for a broad applicability of the model to downstream tasks with datasets of moderate size (for the respective dataset details, please see Table \ref{s-tab:dd}).  To test whether matching all modalities improves performance, we trained a version of OneProt-4 on a reduced pre-training set of 300K datapoints, ensuring every datapoint had all four modalities. While this on average increased modality alignment (see Table \ref{s-tab:ablations}; OneProt-4 matched), the downstream results were worse than for OneProt-4, Table \ref{s-tab:downstream}. Thus, we conclude that modality alignment alone does not determine downstream performance, as the dataset size also plays an important role, even when modalities are not fully matched. Therefore, OneProt is a lightweight, versatile model that efficiently utilizes information stored across modalities without pairing them directly. %This enriched alignment is reflected in \OneProt’s superior performance relative to ESM-2. 

%In the thermostability task, \OneProt’s performance lags behind other models, potentially due to the unique demands of regression tasks in comparison to classification tasks. %Overall, \OneProt’s performance underscores its robust design, effectively integrating information across modalities to support a range of protein function predictions.
%[XXX: OneProt is less strong on the Thermostab. task, which is a regression task(?). Discuss this?]

\subsection*{Enzyme function prediction}\label{subsec:topenzyme}

\begin{figure}[!h]%[!htpb]
    \centering
    \includegraphics[width=0.8\linewidth]{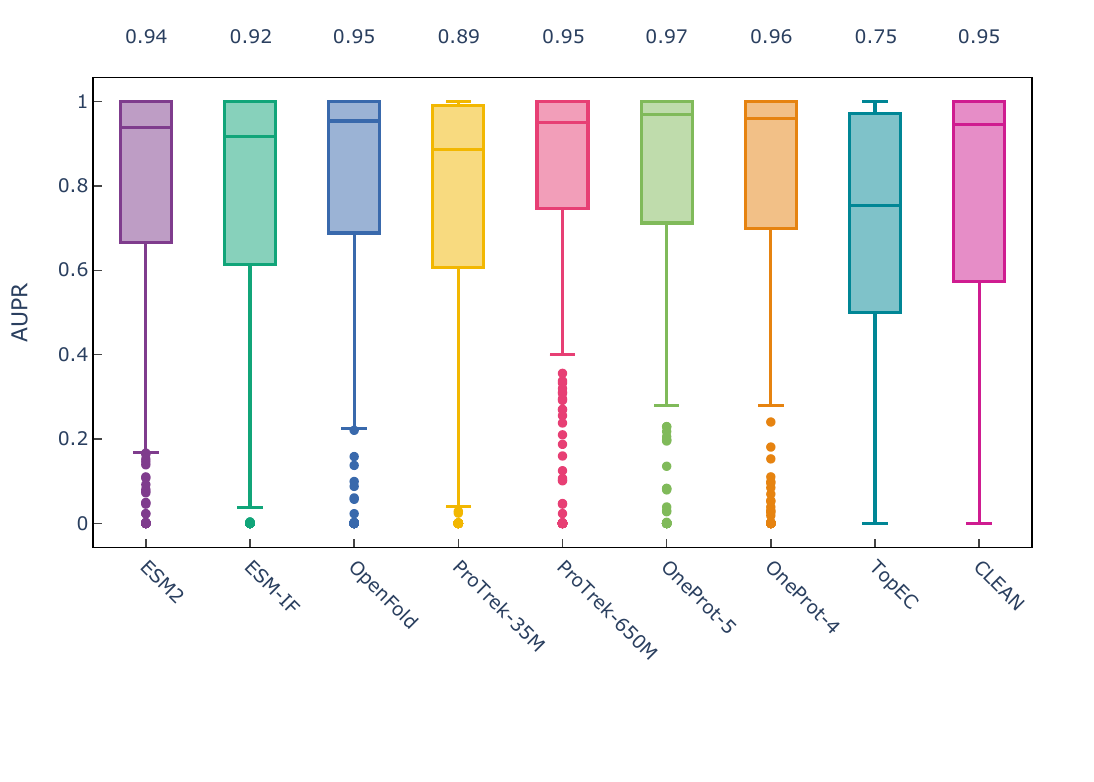}
    \caption{{\bf Model performance comparison based on Area Under Precision Recall curve (AUPR) scores for TopEnzyme.}  Each boxplot shows the AUPR distribution for a method (TopEC, CLEAN, ESM-2, Protrek-35M, ProTrek-650M, OneProt).  %\\
    %{\bf Alt Text:} Box plots displaying the AUPR distributions of the TopEnzyme EC numbers prediction across 6 methods with median AUPR value stated above each plot.
    }
    \label{fig:topenzyme_boxplot}
\end{figure}

Results of the analysis described in the \nameref{oneprotmodel} are presented in Fig.~\ref{fig:topenzyme_boxplot}. All embedding-based methods, ESM, ProTrek, and OneProt outperform  TopEC  ($p<10^{-6}$), while only OneProt-5 significantly outperforms CLEAN ($p<0.04$). We note, however, that CLEAN results in a higher IQR ($0.43$) than any one of the embedding-based methods ($0.25-0.38$), as reported in Table \ref{s-tab:topenzyme_outliers}. %Interestingly, the MLP based on ESM2 embeddings is better than CLEAN, which is a contrastive learning method built on top of ESM2. 
 OneProt-5 achieves significantly higher median AUPR values than the ESM, OpenFold, and ProTrek models ($p<0.03$). Similarly, OneProt-4 significantly outperforms the ESM and ProTrek-35M models ($p<0.02$), while delivering a performance comparable to OpenFold and ProTrek-650M, with a tendency toward higher values ($p>0.2$). Note, on this markedly larger dataset compared to the previous section \nameref{subsec: saprot tasks}, the OneProt-5 model with the higher number of modalities  shows a trend towards better results than OneProt-4 with a tighter  IQR. Moreover, as summarized in Table \ref{s-tab:topenzyme_outliers}, OneProt-5 also has a low number of outliers (51), corresponding to the values beyond $Q_1-1.5$IQR.  The class size distribution of the outliers results in significantly lower values, compared to the overall class size distribution (one-sided Wilcoxon rank sum test $p<10^{-6}$). Therefore, we assume that extending the underrepresented classes could potentially improve the learning and lead to better classification results. We note that CLEAN and TopEC do not have any outliers at all, but they also result in significantly poorer AUPR values than all other models,  such that the  aforementioned equation yields outlier thresholds with negative values.

The increased performance of ProTrek-650M and OneProt over single-modality methods such as TopEC, CLEAN, and ESM-2 shows the potential of aligned multi-modal protein representations for downstream applications in structural biology. 

 % \begin{table}[!htbp]
 % \centering
 % \caption{ Results for Predicting EC Numbers at the Full Hierarchy}
 % \label{tab:topenzyme-results}
 % \begin{tabular}{@{}ll@{}}\toprule
 % Model & Accuracy \\ \midrule
 % TopEC & $72.$  \\ 
 % CLEAN & $74.$ \\
 % ESM-2 (ours) & $78.6$ \\ 
 % \OneProt (ours) & $87.9$ \\ \midrule
 % \end{tabular}
 % \end{table}

 % \begin{greenbox}{}
 % \textbf{Take-aways:} \OneProt demonstrates a strong performance on the challenging TopEnzyme task, achieving 87.9\% accuracy in enzyme function prediction. By generating embeddings for use with simple MLP classifiers, \OneProt highlights the effectiveness of aligned multi-modal protein representations for complex classification tasks in structural biology.
 % \end{greenbox}

\subsection*{Representation learning for evolutionarily related proteins}
\label{subsec: Representation Learning}

\begin{figure*}[!h]
    \centering
    \includegraphics[width=0.95\textwidth]{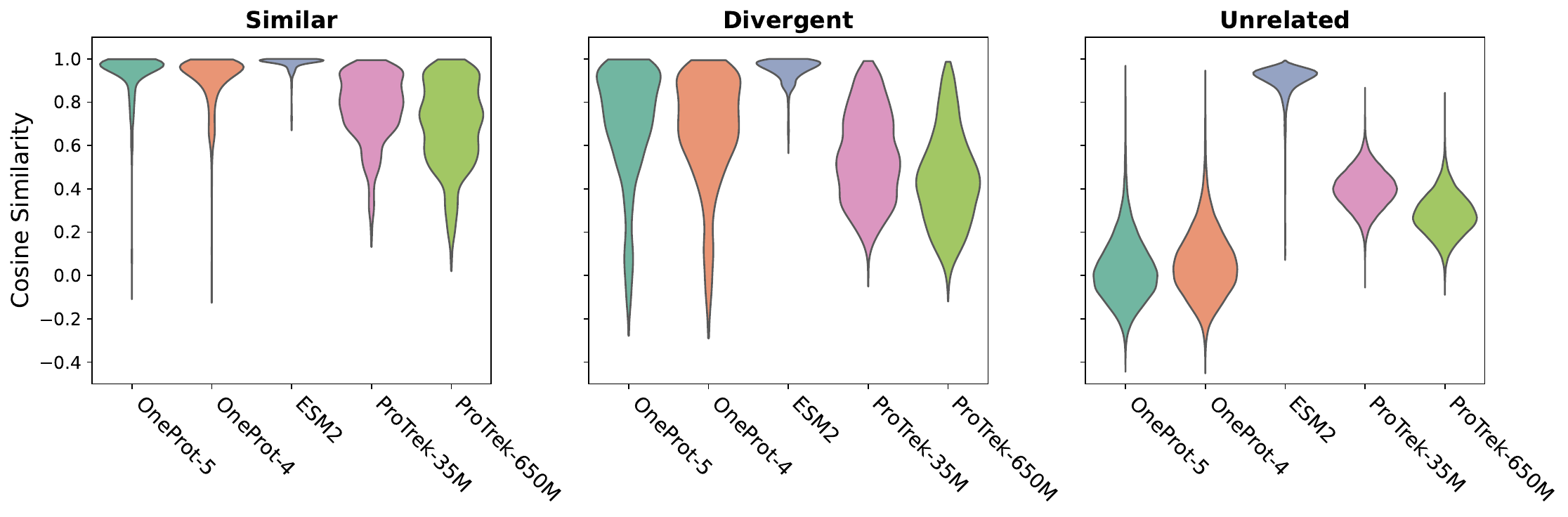}
    \caption{{\bf Cosine Similarity distributions for models ESM-2, ProTrek-35M and -650M, OneProt-4 and -5.}
The plot shows the similarity of a given protein to three groups: the 50 most evolutionarily similar proteins, the 50 most evolutionarily divergent sequences, and 1000 unrelated sequences. While all models partially capture evolutionary relationships, OneProt distinctly separates the three classes, demonstrating its ability to generate meaningful sequence representations.%\\
%{\bf Alt Text:} Three violin plots illustrating cosine similarity distributions among ESM2, ProTrek and OneProt models across similar, divergent and unrelated sequences.
}
    \label{fig:similarity_metrics_revised}
\end{figure*}

Fig~\ref{fig:similarity_metrics_revised} presents violin plots of cosine similarities  across three sequence categories, evolutionarily related similar sequences, evolutionarily related divergent sequences and evolutionarily unrelated sequences, as described in the \nameref{oneprotmodel}, for the selected models. 

%We extend our evaluation by introducing a new baseline, the ProTrek \cite{protrek} sequence encoder, a 650M model following the ImageBind paradigm, for a methodological comparison. 
Embeddings from both OneProt models outperform baselines in capturing evolutionary relationships, indicated by a high cosine similarity among related sequences (Fig~\ref{fig:similarity_metrics_revised} left and middle),
whereas a significantly lower similarity with a large margin for separation is found for unrelated sequences (Fig~\ref{fig:similarity_metrics_revised} right, $p<10^{-16}$). We note that ESM-2 and OneProt generally yield high cosine similarity values across related sequence types ( $p<10^{-16}$ compared to ProTrek), while ProTrek models show less distinction between dissimilar and unrelated sequences ( separation margins $<0.2$ for ProTrek, $>0.6$ for OneProt, $p<10^{-16}$). OneProt's superior capacity to distinguish homologous from non-homologous sequences is likely due to OneProt's %[XXX: Just above we say that OneProt is better than ESM2 and ProTrek. Here, we consider ProTrek and OneProt together, which goes against "this distinction". Delete "Protrek's and "?] 
multi-modal training with the InfoNCE loss (Eq (\ref{eq: infonce loss})). This facilitates the exchange of representational information between encoders, resulting in enriched embeddings that are capable of distinguishing evolutionary relationships. Since OneProt models do not encounter MSA data during training directly, the ability to separate related and unrelated sequences reflects the information shared across encoders. %Investigating which encoder contributes this information could provide valuable insights into OneProt's representational strengths.

\subsection*{ProSPECCTs}
\label{subsec: prosspects}

We present the AUC results of the analysis as shown in Table \ref{tab:prospecct_auc}. The last three columns correspond to the OneProt ablations, which comprise different encoders in addition to the sequence one: structure graph and text encoders only (SG+Text), structure token and text encoders only (ST+Text), structure token, pocket, and text (ST+Text+Pocket). These results showcase the importance of the pocket modality and are discussed in the following section, \nameref{subsec:ablations}.%The respective plots are available as a supplementary figure \ref{s-fig:ROC_curves_prospeccts}. 
\begin{table*}[!h]
\centering

    \renewcommand{\arraystretch}{1.0} 
    
    % Push caption and table to the left by removing centering and using flushleft
  \begin{adjustwidth}{-1.0in}{0in}
\caption{AUC Scores for the ProSPECCTs datasets. ST and SG stand for Structure Token and Structure Graph modalities, respectively.}%For the last two columns concatenated embeddings are used for prediction.}
\label{tab:prospecct_auc}
\begin{tabular}{l*{8}{c}}  % Changed to center alignment for numbers
\toprule
\multirow{2}{*}{\textbf{Dataset}} & 
\multicolumn{3}{c}{\textbf{Base Models}} & 
\multicolumn{5}{c}{\textbf{OneProt Models}} \\  % Changed to 5 columns
\cmidrule(lr){2-4} \cmidrule(lr){5-9}  % Extended to column 9
& ESM-2 & ProTrek & ProTrek & OneProt-5 & OneProt-4 & SG+Text  & ST+Text & ST+Text \\
& & 650M & 35M &  &  &  &   & +Pocket \\
\midrule
DS1 & 1.000 & 1.000 & 1.000 & 1.000 & 1.000 & 1.000 & 1.000 & 1.000 \\
DS1.2 & 1.000 & 1.000 & 1.000 & 1.000 & 1.000 & 1.000 & 1.000 & 1.000 \\
DS2 & 1.000 & 1.000 & 1.000 & 1.000 & 1.000 & 0.998 & 0.998 & 1.000 \\
DS3 & 0.878 & 0.577 & 0.578 & 0.881 & 0.874 & 0.801 & 0.868 & \textbf{0.899} \\
DS4 & 0.866 & 0.576 & 0.581 & 0.859 &0.850 & 0.779 & 0.877 & \textbf{0.880}  \\
DS5 \& 5.2 & 0.529 & 0.578 & 0.585 & 0.646 & 0.632 &0.639 & \textbf{0.653} & 0.650  \\
DS6 & 0.520 & 0.593 & 0.531 & \textbf{0.621} & 0.551 & 0.527 & 0.602 & 0.587 \\
DS6.2 & 0.520 & 0.592 & 0.531 & \textbf{0.620} & 0.550 & 0.527 & 0.602 & 0.587 \\
DS7 & 0.654 & 0.782 & 0.703 & 0.843 & \textbf{0.848} & 0.834 & 0.840 & 0.843 \\
\bottomrule
\end{tabular}
%}
\end{adjustwidth}
\end{table*}

%We note that all models exhibit excellent performance on DS1-2 and, therefore, do not provide [XXX: What means "do not provide"?] the corresponding ROC curves.
Remarkably,  even without task-specific fine-tuning, OneProt models  tend to outperform all others across all datasets. Notably, for DS5-7, all OneProt models achieve  substantially higher AUC values compared to the baselines.

Given that OneProt-5 exhibited consistently good performance across all ProSPECCTs datasets, surpassing ESM-2 everywhere but DS4, we have also investigated the predictive power of its structure, pocket, and joint concatenated embeddings of OneProt-5 (see Table \ref{s-tab:prospecct_auc_emb}). Yet, the OneProt-5 sequence encoder typically remained superior.
Namely, despite the pocket encoder being trained on binding site data, the sequence encoder in OneProt-5 outperforms it, likely due to its larger pre-training dataset providing a broader understanding of protein properties.  The one exception is DS7, where pocket embeddings and combined structure–sequence embeddings surpass the OneProt sequence embeddings. We attribute this to DS7’s focus on recovering binding-site similarity across highly divergent proteins, making structural and pocket features especially discriminative in that setting.

By integrating sequence-level knowledge with binding site-specific data during pre-training, OneProt demonstrates an enhanced ability to distinguish subtle chemical variations, as systematically evaluated in diverse contexts using the ProSPECCTs benchmark datasets.  Moreover, it showcases the added benefit of multi-modal training atop of the pre-trained ESM-2. 

OneProt's representations appear to better understand the subtle molecular recognition distinctions between protein-ligand complexes, particularly in datasets where proteins bind to the same ligand but exhibit slight structural variations. This observation is especially pronounced in DS3 and DS4, where protein structures are generally very similar, but nuanced changes in chemical interactions are present. %We note also that the models without pocket modality generally perform worse than those including it, suggesting that our pocket representation may capture the critical information required to discern these minute chemical modifications. 

\subsection*{Ablations}\label{subsec:ablations}

We discuss here  the most noteworthy ablations of OneProt, while the  data for the rest, including the Figures, are provided in Tables \ref{s-tab:ablations} - \ref{s-tab:prospecct_auc_emb}, Figures \ref{s-fig:heatmaps_p} - \ref{s-fig:topenzyme_boxplot}. 

\begin{figure}[htbp]
 
\center
   \includegraphics[width=1.0\textwidth]{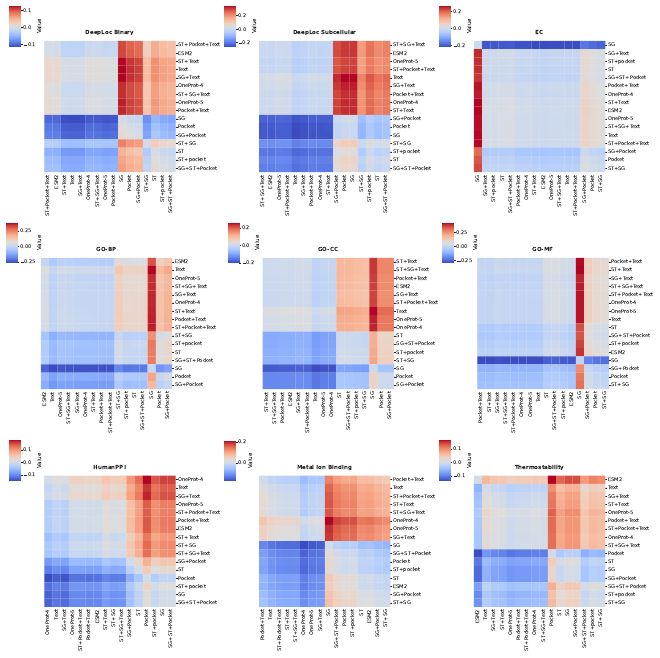}
    \caption{{\bf Normalized performance drops $\Delta_{x,y}=\frac{Perf_y-Perf_x}{Perf_x}$.} The heatmaps visualize the Eq (\ref{eq:norm_drops}) applied to the OneProt ablations, where $x$ is the model on the horizontal axis, while $y$ is the model on the vertical axis. Colours in the shades of blue correspond to negative values (model on the horizontal axis outperforms model on the vertical axis), and colours in the shades of red correspond to positive ones.}
    \label{fig:norm_performance}
\end{figure}

Ablations corresponding to section \nameref{subsec: saprot tasks} are summarized in Table \ref{tab:ablations}, \ref{s-tab:downstream}  and Fig \ref{fig:norm_performance}. The latter presents a heatmap, where negative $\Delta_{x,y}$, Eq (\ref{eq:norm_drops}), indicating inferiority of the model on the vertical axis $y$ compared to model on the horizontal axis $x$, are shown in shades of blue, and positive $\Delta_{x,y}$ in shades of red.  We note that the model comprising only sequence and text encoders exhibits outstanding performance on DeepLoc tasks. We attribute the high accuracy on the  DeepLoc2 downstream task to the alignment between sequence and text encoders: all models, where the sequence-text alignment was higher than 0.22, exhibited an accuracy above 92.7\%,  as shown in Fig.~\ref{fig:norm_performance} (DeepLoc Binary), where three darkest vertical blue lines mark the text, text and structure graph (Text+SG), text and structure token (Text+ST) models.  Moreover, the corresponding three models also exhibited statistically comparable performance with ProTrek-35M ($p>0.09$, two-sided test). This, however, does not guarantee high performance on the  DeepLoc10 task, where a  relatively high 82.2\% accuracy is achieved also by the encoder comprising sequence, structure graph, and text (SG+Text). The latter may have to do with the fact that the structure modality in that case is almost perfectly aligned with the sequence (R@1$>$0.95, R@10=1.0), as well as with the satisfactory emergent structure graph-text alignment (R@1$\approx$0.1, R@10$\approx$0.4), while the text-sequence alignment remains at a high level (Table \ref{s-tab:ablations}).  Clustering in Fig. \ref{fig:norm_performance} indicates the best performing models as those comprising the text encoder (Text), with Text and SG+Text corresponding to the darkest vertical blue lines on the heatmap.

\begin{table*}[!h]
    \small % Make font smaller (you can also use \footnotesize or \tiny for even smaller)
    \renewcommand{\arraystretch}{1.0} 
    
    % Push caption and table to the left by removing centering and using flushleft
    \begin{adjustwidth}{-1.5in}{0in}
    \caption{Downstream results of the selected ablations on the datasets from \cite{Saprot}. Abbreviations as in Tables \ref{tab: saprot tasks}-\ref{tab:prospecct_auc}.}
    \label{tab:ablations}
    % Resize table to fit page width - adjust the scale factor as needed
    \resizebox{1.55\textwidth}{!}{%
    \begin{tabular}{lccccccccc}
        \hline
        \multirow{3}{*}{\textbf{Model}} & \multirow{2}{*}{\textbf{Thermostability}} & \multirow{2}{*}{\textbf{HumanPPI}} & \multirow{2}{*}{\textbf{Metal Ion Binding}} & \multirow{2}{*}{\textbf{EC}} & \multicolumn{3}{c}{\textbf{GO}} & \multicolumn{2}{c}{\textbf{DeepLoc}} \\  
        \cline{6-10} 
                    &  &    &         &  & MF   & BP   & CC   & Subcellular & Binary \\
                    & Spearman's $\rho$ & ACC/AUC\%    & ACC/AUC\%        & Fmax & Fmax   & Fmax   & Fmax   & ACC\% & ACC/AUC\% \\ 
        \hline
        \textbf{Text}            & 0.656 (0.005)        & 87.5/94.3 (1.4/0.5)          & 74.0/81.3 (2.2/2.0)          & 0.876 (0.003)        & 0.656 (0.004)        & \textbf{0.503} (0.002) & \textbf{0.561} (0.005) & \textbf{83.0} (0.4)    & \textbf{92.9/97.3} (0.2/0.1) \\

\textbf{Text+SG}        & 0.664 (0.015) & \textbf{88.1/96.4} (1.3/1.3) & \textbf{75.9/85.1} (0.8/0.4) & 0.866 (0.003) & 0.651 (0.004) & 0.495 (0.004) & 0.545 (0.004) & 82.2 (0.3) & \textbf{92.9/96.9} (0.4/0.3) \\

\textbf{Text+ST}        & 0.666 (0.014) & 84.5/92.7 (1.2/1.1) & 74.7/81.8 (1.1/1.7) & \textbf{0.877} (0.001)   & \textbf{0.661} (0.002) & 0.496 (0.003) & 0.549 (0.007) & 81.6 (0.4) & 92.8/96.9 (0.3/0.1) \\

\textbf{Text+ST+Pocket} & \textbf{0.670} (0.006) & 85.9/93.9 (1.6/0.5) & 74.5/81.8 (1.0/0.5) & 0.876 (0.004) & 0.657 (0.005) & 0.497 (0.004) & 0.546 (0.004) & 80.5 (0.5) & 91.3/96.3 (0.3/0.2)\\
       \hline
    \end{tabular}%
    }
    \end{adjustwidth}
\end{table*}

We observe that combining the structure graph and pocket encoders significantly improves performance on the HumanPPI and Metal Ion Binding tasks ($p<0.04$ and $p<0.003$, respectively, comparing SG+Pocket+Text to a Text-only model, one-sided tests). This improvement is likely due to the complementary information that these modalities provide about binding and interaction sites. Replacing the structure graph with the structure token  (ST) encoder, however, reduces performance ($p<10^{-3}$ comparing SG+Pocket+Text with ST+Pocket+Text, one-sided tests), as the token encoder tends to capture global structural similarity rather than local spatial relationships or graph-like geometries. These aspects are critical for describing interface regions in the HumanPPI task and ensuring local geometric precision in Metal Ion Binding sites.  The superiority of the discussed models is further evident in Fig~\ref{fig:norm_performance}, where the OneProt-4 model is marked by vertical lines in the darkest blue.

For the EC number prediction task, however, we observe that incorporating the ST encoder alongside the text encoder leads to superior performance compared to using the GNN encoder for structure, which reduces the score for this task  ($p < 0.01$ when comparing any model incorporating both ST and Text with models that include SG but omit ST). On this dataset, the information captured by the structure token encoder proves more valuable than the geometric details encoded by the GNN.
In the TopEnzyme task, which also focuses on EC number prediction, the inclusion of structure token and text encoders in the OneProt models results in AUPR medians of at least  0.95, with a  fairly tight distribution of AUPR values (IQR 0.27-0.3) (Figure \ref{s-fig:topenzyme_boxplot}, Table \ref{s-tab:topenzyme_outliers}).  Yet, it is the pocket encoder, which consistently contributes to performance improvements across models. The improvement is statistically significant when comparing ST with ST+Pocket, and ST+SG with ST+SG+Pocket ($p<0.02$). For other model pairs, the pocket encoder generally tends to reduce variability by narrowing the IQR and/or decreasing the number of outliers, while maintaining or slightly improving the median AUPR. An exception is observed in the OneProt-5 model, where the number of outliers did not decrease relative to the ST+SG+Text model, but the median AUPR increased (0.97 vs. 0.95). This emphasizes the importance of adding localized structural information, thereby, confirming the findings of \cite{vanderWeg2024}.  We note that although both EC and TopEnzyme tasks are related to EC numbers predictions, the key encoders for good performance differ, due to the differences in dataset composition. While the TopEnzyme benchmark was specifically designed to better capture EC number differences that depend on active-site geometry, therefore favoring models with pocket-based encoders, the EC task uses broader EC class distinctions and is rich in evolutionary or sequence motif signals. This provides a rationale for why models using ST encoders, which are known to map sequence context well into token space, perform more strongly on the EC task than the TopEnzyme task.

For predicting evolutionarily-related sequences, the model with only the text encoder appears to fail completely, likely because the text encoder focuses on function-to-sequence associations rather than direct sequence comparisons, Figure \ref{s-fig:similarity}. Models relying solely on GNN encoders generally capture evolutionary relationships but exhibit higher variance due to their sensitivity to small structural perturbations. Notably, the combination of GNN structure, pocket, and text encoders achieves superior performance, highlighting the complementary nature of these encoders.

Finally, in the ProSPECCTs task, we note again the importance of the pocket encoder for the DS3-4 datasets, both alone and in combination with other modalities, Table \ref{tab:prospecct_auc}, \ref{s-tab:prospecct_auc}. The predictions on datasets DS6-6.2, on the contrary, significantly benefit from the inclusion of the structure token encoder. The reason, again, lies in the way the datasets were constructed: DS3 and 4 are datasets with similar protein structures but different physicochemical binding sites and binding shape properties, respectively, which can be captured by node and edge features of a GNN pocket encoder; DS6-6.2 are about distant relationships between protein binding sites with identical ligands that have similar environments, which are tasks more suited for the structure token encoder, capturing global structure-based similarities or environment-level context, whereas GNN encoders are more focused on the local information.

In summary, while certain encoders are well-suited to specific downstream tasks, no single encoder model consistently delivers superior performance across the full spectrum of tasks discussed. Moreover, adding certain modalities to a model can degrade performance if the encoder is not well-suited for the requirements of the downstream task. Leveraging the complementary strengths of multiple encoders, while keeping an eye on the modality alignment, is often essential for addressing the diverse and complex nature of protein-related problems. Notably, we highlight the critical role of the GNN structure and pocket encoders, features absent in the otherwise conceptually similar ProTrek models, compared to relying solely on the structure token encoder.

\section{Discussion}
\label{discussion}

This work contributes significantly to the development of multi-modal protein foundation models by implementing a comprehensive framework that extends the ImageBind concept into the protein domain. In addition, we present a lightweight fine-tuning scheme for downstream tasks, designed to align more closely with real-world requirements, enabling efficient adaptation without the need for extensive computational resources or large-scale datasets. The modular and extendable codebase allows easy integration of new modalities using pre-trained encoders, providing a flexible interface for a multitude of downstream tasks, thereby making it highly adaptable and versatile. This design supports efficient ablation studies to assess the impact of specific encoders or their combinations. Based on these insights, we identified two OneProt models, utilizing 4 and 5 encoders, that consistently delivered strong performance across all downstream tasks discussed.  
More specifically, the present study demonstrates emergent alignment between modalities not paired during training, underscoring the framework's capability for cross-modal alignment. Our models yield competitive performance across downstream tasks, including enzyme function prediction, the ProSPECCTs benchmarking initiative, and standard tasks proposed in \cite{Saprot}, often achieving state-of-the-art or better results. 
%The model should be an important building block towards a potential multi-model Protein Foundation Model. [XXX: Ok? What would be missing to reach this goal?] 

Consistently high R@1 values in Fig~\ref{fig:spider_tables} demonstrate OneProt's strong alignment accuracy between trained modality pairs. Despite lacking established benchmarks to quantitatively contextualize these rank metrics, achieving median ranks below 100 (often below 10) across 4,000 samples indicates effective latent space alignment and, thereby, supports the validity of our multi-modal approach. Furthermore, retrieval tasks confirm OneProt’s precise cross-modal alignment, underscoring the robustness and adaptability of our framework in integrating diverse protein data and mirroring ImageBind's achievements \cite{girdhar2023imagebind}, now in the protein domain.

%The outcomes underscore the effectiveness of OneProt’s design in capturing complementary information from diverse modalities, which is crucial for representing the complex, high-dimensional structures-functions-properties [XXX: Ok? Or which landscape is meant? The energy landscape we do not address here (no conformations of the same port. considered).] landscape of proteins.

By generating embeddings for simple MLP classifiers, OneProt highlights the effectiveness of aligned multi-modal protein representations for complex classification tasks in structural biology. As such, OneProt leads to a strong performance on the challenging TopEnzyme task, achieving a median 0.97 AUPR in enzyme function prediction. Moreover, the evaluation on ProSPECCTs benchmark datasets for binding sites analysis demonstrates that our model's sequence embeddings perform consistently better than ESM-2 and ProTrek across all datasets, with the exception of DS4, where OneProt performs comparably to ESM-2, while still beating ProTrek by a pronounced margin, indicating that the additional pocket and structure encoders of OneProt contribute to the results. %Furthermore, as shown in Table \ref{tab:prospecct_auc}, the performance of simple concatenated embeddings generally lags behind. Future research could explore alternative methods for integrating embeddings from multi-modal encoders, as concatenation may not fully capture the complementary information between encoders.
%Adding structure and pocket modalities resulted in more nuanced distinctions, particularly in capturing ligand-induced conformational changes, flexibility, physicochemical properties, and binding site geometries. 
Finally, we demonstrated that OneProt's contrastive learning approach captures evolutionary relationships with greater fidelity than ESM-2 and ProTrek, as evidenced by the lower similarity scores for unrelated sequences. This highlights the efficacy of our multi-modal CLIP training in generating biologically meaningful protein representations.

Importantly, OneProt should not be seen as a direct competitor to more complex multi-modal models, such as SaProt-LoRa \cite{Saprot} or ProTrek-650M \cite{protrek} as shown in Section \nameref{experiments}, where these models retain higher performance in some benchmarks.  Notably, the inclusion of a GNN encoder in OneProt equips it to capture relational dependencies in protein sequences. This design enables OneProt to deliver competitive performance compared to these substantially larger models on tasks including HumanPPI, Metal Ion Binding, TopEnzyme, ProSPECCTs, and evolutionary relatedness, despite being trained on much smaller datasets, with significantly fewer optimizer steps, and using a more memory-efficient fine-tuning scheme compared to SaProt. We also observed that ESM-3 \cite{ESM3}, when applied to the Saprot \cite{Saprot} datasets, did not demonstrate notable improvements in predictive performance. Additionally, it required substantially higher memory during inference and showed reduced effectiveness on the other tasks when sequence length cut-offs were applied. As a result, we opted not to include it in our analysis  beyond the section \nameref{subsec: saprot tasks} to maintain focus on the models that are more computationally efficient and better suited for the specific constraints of our downstream tasks. With this in mind, we note that OneProt demonstrates how aligning specialized encoders can create a cohesive multi-modal system that transfers information effectively across modalities, remaining flexible to incomplete or heterogeneous data while requiring only moderate computational resources. Additionally, the multi-modal training approach has generated representations that robustly preserve evolutionary biological structures, providing a foundation for capturing the complex relationships inherent in protein data across diverse tasks. The models' ablations described in section   \nameref{subsec:ablations} highlight and make understandable the contributions of the specific encoders to downstream tasks. That said, we note that as we add more modalities, our current simple parallelization scheme may become less efficient due to larger memory requirements. In future work, one could adopt more GPU-efficient strategies, which could also potentially handle longer protein sequences. Alternatively, to overcome the memory issues, one could focus on a curated subset of the most relevant modalities, while still using the existing framework.

%The ability to align multiple modalities with protein sequences, even in low-homology [XXX: Does this refer to the 50 proc. cutoff? To me, this is not "low homology". Also, consider this calling "low-identity" (see my comment above)?] scenarios, is an impressive strength of OneProt. This feature is highly relevant for applications in protein engineering, drug discovery, and precision medicine, where understanding nuanced protein characteristics and interactions can lead to innovative therapeutic solutions. For instance, low-homology [XXX: Dito] alignment could facilitate identifying distant protein analogs with similar functions, assisting in the discovery of novel protein functionalities that are otherwise challenging to deduce. Additionally, the model’s integration of advanced modules such as ESM2, ProNet, and MSR BiomedBERT enables it to learn from complex, high-dimensional data, enriching its capacity to analyze and interpret protein-related information. 

%Our experiments validate that constructing a multi-modal framework using aligned protein transformers and graph models is feasible and highly effective in capturing intricate protein data across various representations. Extensive ablation studies underscore the importance of each modality, revealing that a holistic, integrated approach yields superior outcomes. By incorporating these diverse modalities, OneProt sets a promising foundation [XXX: basis? reference?] for the future of multi-modal protein representation, paving the way for impactful applications in various biological research areas.

Beyond current applications, OneProt provides foundational work for expanding multi-modal protein representation models into broader biological tasks, such as multi-target drug design or protein-protein interaction prediction in complex diseases, by integrating OneProt embeddings into diffusion or large language models. This flexibility highlights OneProt’s role in integrative biological research and computational drug discovery, where multi-modal, data-rich environments are essential for achieving translational success. Future research can explore additional modalities, e.g., multiple conformations for a protein and structures of protein-small molecule complexes, protein representations enriched with protein-protein interaction information, using recent encoders \cite{Su2025}, \cite{11059286}, or protein-nucleic acid complexes, acknowledging that such data may not be universally available for all proteins. In addition, information on proteins under different environmental, experimental, or physiological conditions could be added, which may further refine OneProt’s capacity to analyze proteins under various physicochemical conditions.  A deeper analysis of the evolutionary relatedness between embeddings could be conducted, quantifying contributions of structural or protein-protein interaction information, and of the transfer of information from different modalities to the sequence embeddings. Moreover, extending the current contrastive loss framework could allow moving beyond treating only pairs of modalities as positives. Instead, one could emphasize the clustering of proteins that participate in, e.g., the same disease pathway, and incorporate such relational information into the model via the encoders described in \cite{ZHAO2025121360}.

\section*{Data and Code Availability Statement}
The datasets used for model training and downstream tasks are available at the sources described in the respective references. The code repository for OneProt models is available at: \href{https://github.com/klemens-floege/oneprot.git}{https://github.com/klemens-floege/oneprot.git}. The respective data is available at: \href{https://zenodo.org/records/15429594}{https://zenodo.org/records/15429594}.

\section*{Acknowledgements}

AB, EM, SK, MP are supported by the Helmholtz Association Initiative and Networking Fund in the frame of Helmholtz AI. AB, HG, SK are supported by the Helmholtz Foundation Model Initiative within the project “PROFOUND”. HG, KW are supported by the John von Neumann Institute for Computing (NIC) on the supercomputer JUWELS-BOOSTER at JSC (user ID: VSK33, FOUND). VF is supported by a Branco Weiss Fellowship. The funders had no role in study design, data collection and analysis, decision to publish, or preparation of the manuscript.
\section*{Acknowledgement}

This work is supported by the Helmholtz Association Initiative and Networking Fund in the frame of Helmholtz AI as well as by the Helmholtz Foundation Model Initiative within the project “PROFOUND”. We thank the Jülich Supercomputing Centre (JSC) for their support and advice. We are grateful for the computing time provided by the John von Neumann Institute for Computing (NIC) on the supercomputer JUWELS-BOOSTER at JSC (user ID: VSK33, FOUND). VF was supported by a Branco Weiss Fellowship. 

%\end{acknowledgement}

%%%%%%%%%%%%%%%%%%%%%%%%%%%%%%%%%%%%%%%%%%%%%%%%%%%%%%%%%%%%%%%%%%%%%
%% The same is true for Supporting Information, which should use the
%% suppinfo environment.
%%%%%%%%%%%%%%%%%%%%%%%%%%%%%%%%%%%%%%%%%%%%%%%%%%%%%%%%%%%%%%%%%%%%%
%\begin{suppinfo}
%
%Additional training and model details, dataset descriptions, ablation results, summarized in text, tables and figures.
%
%\end{suppinfo}

%%%%%%%%%%%%%%%%%%%%%%%%%%%%%%%%%%%%%%%%%%%%%%%%%%%%%%%%%%%%%%%%%%%%%
%% The appropriate \bibliography command should be placed here.
%% Notice that the class file automatically sets \bibliographystyle
%% and also names the section correctly.
%%%%%%%%%%%%%%%%%%%%%%%%%%%%%%%%%%%%%%%%%%%%%%%%%%%%%%%%%%%%%%%%%%%%%
%\mciteErrorOnUnknownfalse
\bibliographystyle{plainnat}
%\bibliography{achemso-demo}

%\begin{appendices}

%\appendix
\begin{appendix}[Supplementary Information]
\renewcommand{\thesection}{S\arabic{section}} % A, B, C numbering
\setcounter{table}{0} % Reset table counter
\setcounter{figure}{0} % Reset figure counter
\renewcommand{\thetable}{S\arabic{table}}  
\renewcommand{\thefigure}{S\arabic{figure}}

\section*{Supplementary Material}\label{sec:supp}
%\section{Supplementary Information}
\section{Pre-training}
\label{s-supp:pretraining}
\subsection{Training details}
\label{supp: pretraining1}
We used the AdamW optimizer \citep{adamW} with a constant learning rate schedule 0.001, the temperature parameter $\tau$ in the loss function (Equations (\ref{eq: infonce loss})-(\ref{eq: symmetrical loss}) of the main text) set to $1$. The batch size for each of the modalities was 32. 

During training, data from different modalities were combined using PyTorch Lightning's \textsf{CombinedDataloader} in \textsf{min\_size} mode, ensuring that each epoch's dataset size matched the smallest modality. Therefore, at the beginning of each epoch larger modalities were subsampled accordingly. For validation, \textsf{sequential} mode was used to enable exhaustive evaluation on the validation dataset.  %, and the regularization parameter $\lambda$ equal to $5\times 10^3$. 

 The models were run on JUWELS BOOSTER \citep{booster} on 16 nodes with each compute node comprising four NVIDIA A100 GPUs for 33000 optimizer steps. {Parallelization was implemented using PyTorch’s Distributed Data Parallel scheme, which loads the model on each GPU, thereby minimizing inter-node overhead and resulting in faster training performance. Because the model is loaded onto each GPU, it is light enough to run on a single GPU, and we used 64 GPUs only to accelerate training. The seeds were fixed during pre-training and are available in the corresponding config files in the github repository.
%The batch size was chosen based on XXX.
\subsection{Metrics}\label{supp:metrics}

%\begin{table}[!htbp]
%\centering
%\caption{ Overview of OneProt's different encoders}
%\label{tab:encoder-overview}
%\begin{tabular}{@{}rrrrr@{}}\toprule
%Modality & Model & Pooler & Projection & Parameter Count \\ \midrule
%Sequence & ESM2 & Attention 1D &  MLP & $652$ M \\
%Structure - PDB & ProNet & mean &  linear &  $2.6$ M \\
%Structure - Token & ProNet & CLS &  linear &  $35$ M \\
%%MSA & ESM-1b  & mean &  MLP & $117$ M \\
%Pocket & ProNet & mean &  linear & $47.8$ M \\ 
%Text & MSR BiomedBERT & CLS &  MLP & $110$ M \\ \midrule 
%Total: &  & &  & $964.4$ M \\ 
%%Total: &   & $819.4$ M \\  
%\end{tabular}
%\end{table}

To show the model's retrieval capabilities evolving over training, we report the median rank across two types of retrieval tasks. For retrieval between modalities that were trained together (such as sequence paired with another modality), we calculate the median rank of cosine similarities of the same protein representations, one of which includes the sequence representation, and average them out. %similarity scores across all tasks, showing the retrieval performance between pairs that were explicitly optimized during training. 

For emergent retrieval tasks, which involve pairing modalities not directly trained together, we evaluate the ability of the model to generalize to untrained modality pairs. Similarly to the previous case, we measure the median rank of cosine similarity between representations of the same protein across each pair of modalities, not including the sequence one, and average over the respective emergent retrieval tasks. This approach provides a robust indication of how well the model captures transferable features between previously unlinked modalities.
% \begin{align*}
% MR=\frac14\sum_{i\in \mathcal{M}} Median(S_C^{seq,i})\\
% MR_{em}=\frac1{6}\sum_{\substack{i,j\in\mathcal{M},\\  i\neq j\neq seq}}Median(S_C^{i,j})
% \end{align*}

\section{Supervised fine-tuning Downstream Tasks}
\label{experimentdetails: supervised finetuning}

For our supervised downstream tasks, we initially generate embeddings using the sequence reference and projection head of our OneProt model. These embeddings are created for all proteins in the datasets from \cite{Saprot} and \cite{vanderweg_TopEnzyme}, covering a total of ten datasets. Subsequently, we train a Multi-Layer Perceptron (MLP) on these embeddings. We conducted a comprehensive hyperparameter sweep, exploring a variety of settings:

\begin{itemize}
    \item \textbf{Learning Rates:} 0.001, 0.01
    \item \textbf{Batch Sizes:} 32, 64
    \item \textbf{Maximum Epochs:} 50
    \item \textbf{Hidden Dimensions:} [256], [512, 256]
    \item \textbf{Dropout Rates:} 0.1, 0.25
    \item \textbf{Normalization:} Batch normalization (enabled/disabled), Layer normalization (enabled/disabled)
    \item \textbf{Activation Functions:} "relu", "gelu"
    \item \textbf{Residual Connections:} enabled, disabled
\end{itemize}

These hyperparameter combinations provided a robust evaluation of the MLP model across various biological and biochemical tasks, allowing us to fine-tune the model's performance for each specific task while offering the flexibility to experiment with different models efficiently. The MLP-based approach is simpler and considerably faster, as it eliminates the need for fine-tuning the entire protein encoder model using LoRA, as done for the Saprot\cite{Saprot}. This allows for faster iteration and evaluation cycles, as we can directly use precomputed embeddings and quickly train the MLP with various configurations. Moreover, this setup is highly flexible, enabling the easy integration of other supervised learning models, such as logistic regression, random forests, gradient-boosted trees, and support vector machines. This adaptability makes it straightforward to experiment with multiple models and identify the best-performing one for each specific task.

%\subsection{Protein Pockets dataset}
%\label{experimentdetails: pockets}

\section{Dataset Details}
\label{appendix_datasets}

To create the full training dataset, we combined data from multiple sources. We started with the OpenProteinSet \citep{ahdritz2023openproteinset} as a basis for training due to the high availability of MSA data. This database contains structures, sequences, and MSAs for proteins from the PDB \citep{burley2023rcsb} and proteins from UniClust30 \citep{steinegger_uniclust}. We extend this database with proteins from UniProtKB/Swiss-Prot \citep{Boutet2007} as these proteins are experimentally studied and usually have information across multiple modalities available. Using MMseqs2 \citep{steinegger2017mmseqs2}, we clustered the obtained sequences from OpenFold and Swiss-Prot with a sequence identity cut-off of 50\%, such that each cluster represents a homologous cluster in the protein fold space \citep{vanderweg_TopEnzyme}. We align the training, validation, and test split along these sequence clusters. For each cluster representative and member, using the sequence, we find the structure from the AlphaFold2DB \citep{alphafolddb}, the MSA from the OpenProteinSet, and the binding pocket with P2Rank \citep{p2rank} including a 100 closest residues to the predicted binding site, as in \cite{vanderWeg2024}. As we could not find a binding pocket for each protein, fewer data points for these modalities are available.

\section{Abbreviations glossary}

\begin{description}
\item[AI] Artificial Intelligence
  \item[AUC] Area Under Receiver Operatoing Characteristic curve
\item[AUPR] Area Under Precision-Recall curve
\item[BP] Biological Process
\item[CC] Cellular Component
\item[CLIP] Contrastive Language-Image Pre-training
\item[DDP] Distributed Data Parallel
\item[EC] Enzyme Commission
\item[ESM] Evolutionary Scale Modeling
\item[IF] Inverse Fold
\item[IQR] Inter-Quantile Range
\item[GNN] Graph Neural Network
\item[GO] Gene Ontology
\item[GPT] Generative Pre-trained Transformer
\item[GPU] Graphics Processing Unit
\item[HumanPPI] Human Protein Protein Interactions
\item[InfoNCE] Information Noise-Contrastive Estimation
\item[LoRA] Low-rank Adaptation
\item[MF] Molecular Function
\item[MLM] Masked Language Modeling
\item[MLP] Multi-Layer Perceptron
\item[MR] Median Rank
  \item[MSA] Multiple Sequence Alignment
  \item[NMR] Nuclear Magnetic Resonance
  \item[ProSPECCTs] Protein Site Pairs for the Evaluation of Cavity Comparison Tools
  \item[SG] Structure Graph
  \item[ST] Structure Token
\end{description}

\begin{table}[H]
\captionsetup{skip=1pt}
\centering
\caption{Overview of supervised Downstream Datasets from \cite{Saprot} and \cite{vanderweg_TopEnzyme}.  The data splits follow those provided in the original studies and reflect the same clustering strategy used therein. For multi-label classification tasks the number of classes is listed in parentheses.}\label{s-tab:dd}
\makebox[\textwidth][c]{
\resizebox{1.4\linewidth}{!}{%
\begin{tabular}{@{}llllrrr@{}}
\toprule
\textbf{Dataset} & \textbf{Type} & \textbf{Category} & \textbf{Evaluation Metric} & \textbf{Train} & \textbf{Valid} & \textbf{Test} \\ \midrule
Thermostability \citep{Dallago2021FLIP} & Regression & Human-Cell & Spearman's $\rho$ & 5056 & 639 & 1336 \\
MetalIonBinding \citep{Hu2022Exploring} & Binary classification & - & Acc & 4247 & 662 & 665 \\
\multirow{2}{*}{DeepLoc \citep{AlmagroArmenteros2017}} & Binary classification & Binary & Acc & 5477 & 1336 & 1731 \\
 & Multi-class classification (10) & Subcellular & Acc & 8747 & 2191 & 2747 \\
HumanPPI \citep{Xu2022PEER} & Binary classification & - & Acc & 26319 & 234 & 180 \\
EC \citep{Gligorijevic2021Structure} & Multi-label classification (585) & - & Fmax & 13089 & 1465 & 1604 \\
GO \citep{Gligorijevic2021Structure} & Multi-label classification (1943 / 489 / 320) & BP / MF / CC & Fmax & 26224 & 2904 & 3350 \\
%TopEnzyme \cite{vanderweg_TopEnzyme} & Multi-label classification & - & Fmax & 224531 &15149 & 17488\\
\hline
TopEnzyme \cite{vanderweg_TopEnzyme} & Multi-label classification (826) & - & Fmax & 224531 & 15149 & 17488 \\
\bottomrule
\end{tabular}}
}
\end{table}

%\subsection{PROSSPECTS}
\begin{table}[H]
\captionsetup{skip=1pt}
\centering
\caption{Overview of ProSPECCTs Datasets. }
%\label{tab: prospecct dataset}
\makebox[\textwidth][c]{ 
\resizebox{1.3\linewidth}{!}{%
\begin{tabular}{@{}llllrrr@{}}
\toprule
\textbf{Dataset} & \textbf{Number of pockets} & \textbf{Dataset description} \\ \midrule
DS1 & 326 & Structures with identical sequences but different ligands \\
DS1.2 & 45 & Structures with identical sequences and similar ligands \\
DS2 & 329 & Flexible NMR structures \\
DS3 & 1954 & Similar structures with different physicochemical binding site \\
DS4 & 1954 & Similar structures with different binding site shape properties \\
DS5 \& 5.2 & 100 & Similar proteins binding to identical ligands and cofactors, including\\
& & phosphate binding sites (DS5.2) \\
DS6 & 35 & Distant relationships between protein binding sites but identical ligands\\
& & that have a similar environment \\
DS6.2 & 35 & Same as in 6, additionally include cofactors \\
DS7 & 49 & The recovery of known binding sites similarities within a diverse set \\
& & of proteins is tested \\
\bottomrule
\end{tabular}\label{s-tab:prospecct_dataset}
}
}
\end{table}

%\clearpage 
% \begin{figure*}[!h]
%     \centering
%     \includegraphics[width=1.0\textwidth]{figures/roc_plot_v2.pdf}
%     \caption{Receiver Operating Characteristic (ROC) curves for the classifications of ProSPECCTs DS3-7 datasets for different models.}
% %    {\bf Alt Text}: 9 ROC curves representing performance of 5 different \OneProt models, 2 ProTrek models and ESM2 on DS1-7 PRoSPECCTs datasets}
%     \label{fig:ROC_curves_prospeccts}
% \end{figure*}

%\section{Ablations}\label{supp:ablations}

%\subsection{Modality alignment}\label{supp:alignment}
%In all the following figures and tables ST and SG stand for Structure Token and Structure Graph modalities respectively as in Table \ref{m-tab:prospecct_auc} of the main text.
\begin{table*}[!htbp]
\captionsetup{skip=1pt}
\centering
\caption{Modality alignments for selected ablations in terms of R@1 and (in parentheses) R@10.  ST corresponds to Structure Token modality, SG corresponds to Structure Graph modality, '+' indicates a combination of multiple modalities.}
\label{s-tab:ablations}
\small  % Reduced font size for better fit
\makebox[\textwidth][c]{ 
\resizebox{1.15\linewidth}{!}{
\begin{tabular}{l|*{7}{c}}  % Changed to center alignment for numbers
\toprule
%\multirow{2}{*}{\textbf{Algnment}} & \\ 
%\multicolumn{3}{c}{\textbf{Base Models}} & 
%\multicolumn{5}{c}{\textbf{OneProt Models}} \\  % Changed to 5 columns
%\cmidrule(lr){2-4} \cmidrule(lr){5-9}  % Extended to column 9
\textbf{Algnment} & Text & Text+ST & Text+SG & Text+pocket & Text+ST & Text+SG   &OneProt-4\\
& &  &  &  &+pocket  &+ST  & matched \\
\midrule
seq-text & 0.26 (0.68) & 0.24 (0.64) & 0.24 (0.63) & 0.22 (0.63) & 0.20 (0.60) & 0.20 (0.60) & 0.23 (0.62)  \\
text-seq & 0.24 (0.65) & 0.22 (0.62) & 0.20 (0.60)  & 0.20 (0.61) & 0.19 (0.58) & 0.18 (0.58) & 0.25 (0.65) \\
seq-ST & - & 0.45 (0.88) & - & - & 0.46 (0.90)  & 0.48 (0.89) & - \\
ST-seq & - & 0.43 (0.88) & - & - & 0.46 (0.90)& 0.45 (0.89) & - \\
seq-SG & - & - & 0.97 (1.0) & - &-  & 0.84 (1.0) & 0.89 (1.0)  \\
SG-seq  & - & - & 0.95 (1.0) & - &- & 0.84 (1.0) & 0.90 (1.0)\\
seq-pocket & - & - & - & 0.33 & 0.30 (0.74) &-  & 0.38 (0.78)\\
pocket-seq & - & - & - &0.30  & 0.28 (0.72) & -  & 0.35 (0.76)\\
SG-text & - & - & 0.13 (0.47) & - & - & 0.11 (0.41) &  0.14 (0.49)\\
text-SG & - & - & 0.11 (0.43) & - & - & 0.10 (0.37) & 0.13 (0.46)\\
ST-text & - & 0.08 (0.35) & - & - & 0.08 (0.34)& 0.07 (0.31) & -\\
text-ST & - & 0.07 (0.34) & - & - & 0.07 (0.32)& 0.06 (0.29) & -\\
pocket-text & - & - & - & 0.06 (0.28) & 0.05 (0.25) & -  & 0.07 (0.30)\\
text-pocket & - & - & - & 0.07 (0.28) & 0.05 (0.25)& -  & 0.07 (0.30)\\
pocket-SG & - & - & - & - & - & -  & 0.36 (0.78) \\
SG-pocket & - & - & - & - & - & -  & 0.39 (0.80)\\
\bottomrule
\end{tabular}
}}
\end{table*}

%\subsection{Downstream tasks}\label{supp:ablation-down}
\begin{table*}[!t]
\captionsetup{skip=1pt}
\centering

\renewcommand{\arraystretch}{1.2} 
\caption{Downstream results for the ablations not included in the main text on the datasets from \cite{Saprot}. The tasks comprise ThermoStability (regression) evaluated using Spearman correlation, HumanPPI, Metal Ion Binding, DeepLoc Binary (binary classification) and DeepLoc Subcellular (multiclass classification), evaluated using accuracy (ACC) and Area Under the Reciever Operating Curve (AUC), Enzyme Commision numbers (EC), Gene Ontology (GO) terms corresponding to Molecular Function (MF), Biological Process (BP) and Cellular Component (CC) evaluated using maximum F1-score metric (Fmax) defined by formula (\ref{eq:fmax}) of the main text. ST corresponds to Structure Token modality, SG corresponds to Structure Graph modality, '+' indicates a combination of multiple modalities.}\label{s-tab:downstream}%\OneProt demonstrates competitive performance across all evaluated tasks, generally surpassing the ESM-2 \& SaProt baselines and outperforming the stronger SaProt-LoRA model in several key tasks. }

%\begin{tabular}{|l|c|c|c|c|ccc|cc|}
\makebox[\textwidth][c]{ 
\resizebox{1.5\linewidth}{!}{%
\begin{tabular}{lccccccccc}
\hline
\multirow{3}{*}{\textbf{Model}} & \multirow{2}{*}{\textbf{Thermostability}} & \multirow{2}{*}{\textbf{HumanPPI}} & \multirow{2}{*}{\textbf{Metal Ion Binding}} & \multirow{2}{*}{\textbf{EC}} & \multicolumn{3}{c}{\textbf{GO}} & \multicolumn{2}{c}{\textbf{DeepLoc}} \\  \cline{6-10} 
            &  &    &         &  & MF   & BP   & CC   & Subcellular & Binary \\   %\hline
               & Spearman's $\rho$ & ACC\%    & ACC\%        & Fmax & Fmax   & Fmax   & Fmax   & ACC\% & ACC\% \\ 
\hline

\textbf{Pocket}       & 0.601 (0.022) & 75.8/84.8 (1.1/1.2) & 66.1/70.8 (1.4/1.5) & 0.843 (0.003) & 0.604 (0.027) & 0.437 (0.003) & 0.473 (0.003) & 62.1 (0.9) & 83.6 (0.6) \\

\textbf{Text+Pocket}  & \textbf{0.670} (0.008) & \textbf{86.1/93.7} (1.6/0.7) & 72.6/83.0 (0.9/0.7) & 0.872 (0.004) & \textbf{0.659} (0.003) & 0.497 (0.003) & 0.547 (0.006) & \textbf{81.5} (0.4) & \textbf{92.3/96.5} (0.2/0.1) \\

\textbf{SG}           & 0.616 (0.011) & 77.3/84.2 (1.6/1.2) & 63.9/72.3 (1.6/1.9) & 0.698 (0.004) & 0.467 (0.004) & 0.367 (0.003) & 0.447 (0.011) & 62.0 (0.7) & 82.8/89.5 (0.7/0.7) \\

\textbf{SG+Pocket}    & 0.616 (0.011) & 81.2/87.2 (1.1/0.9) & 68.2/70.8 (2.0/0.9) & 0.829 (0.001) & 0.589 (0.005) & 0.425 (0.003) & 0.473 (0.003) & 63.0 (0.3) & 84.0/90.4 (0.6/0.4) \\

\textbf{ST}           & 0.623 (0.018) & 79.1/89.0 (1.7/0.6) & 67.6/78.2 (2.5/0.4) & 0.863 (0.003) & 0.630 (0.005) & 0.459 (0.003) & 0.497 (0.011) & 66.9 (0.7) & 86.8/92.8 (0.4/0.2) \\

\textbf{ST+Pocket}    & 0.633 (0.016) & 78.2/83.8 (2.5/1.1) & 65.9/73.6 (2.0/1.2) & 0.865 (0.003) & 0.637 (0.004) & 0.461 (0.003) & 0.501 (0.009) & 65.2 (0.7) & 87.6/92.1 (0.5/0.3) \\

\textbf{ST+SG}        & 0.636 (0.006) & 85.3/90.9 (1.4/1.8) & 68.6/74.8 (0.8/2.1) & 0.850 (0.003) & 0.609 (0.026) & 0.448 (0.002) & 0.503 (0.008) & 68.1 (0.7) & 89.2/94.2 (0.4/0.2) \\

\textbf{ST+SG+Text}   & 0.669 (0.006) & 85.0/93.1 (1.1/0.6) & \textbf{74.6/83.2} (0.3/1.2) & \textbf{0.875} (0.002) & 0.654 (0.004) & 0.493 (0.002) & \textbf{0.549} (0.003) & 80.9 (0.4) & 92.1/93.1 (0.2/0.2) \\

\textbf{ST+SG+Pocket} & 0.642 (0.005) & 77.3/85.6 (1.9/1.7) & 65.5/71.8 (2.7/1.5) & 0.861 (0.005) & 0.627 (0.003) & 0.459 (0.003) & 0.498 (0.008) & 66.2 (0.5) & 87.9/93.1 (0.5/0.5)  \\
\textbf{OneProt-4 matched} &0.647 (0.011) &85.7/94.5 (0.9/0.3) &74.3/81.8 (1.5/0.9) & 0.867 (0.004) & 0.651 (0.003) & 0.491 (0.002) & 0.549 (0.008) & 80.8 (0.4) &91.9/96.3 (0.4/0.1)\\
%\textbf{ProTrek-650M} & 0.628 & 91.7 & 78.3 & 0.877 & 0.678 & 0.543 & 0.618 & 91.1 & 95.4  \\
%\rowcolor{yellow}
%\textbf{OneProt-5}          & 0.681 & 82.2  & 76.0 & 0.873 & 0.653 & 0.486  & 0.562 & 80.2 & 92.0 \\
%\rowcolor{yellow}
%\textbf{OneProt-4}        & 0.656 & 89.4  & 77.4 & 0.867 & 0.656 & 0.498  & 0.552 & 81.2 & 91.8 \\
\hline
\end{tabular}
}}
\end{table*}
\clearpage

\begin{table}[H]
\centering
\captionsetup{skip=1pt}
\caption{Receiver Operating Characteristic Area Under the Curve (AUC) Scores for the ProSPECCTs datasets for the ablations of OneProt not included in the main text. ST corresponds to Structure Token modality, SG corresponds to Structure Graph modality, '+' indicates a combination of multiple modalities.}
\label{s-tab:prospecct_auc}
\small  % Reduced font size for better fit
\makebox[\textwidth][c]{ 
\resizebox{1.5\linewidth}{!}{
\begin{tabular}{lrrrrrrrrrr}  % Changed to center alignment for numbers
\toprule
\textbf{Dataset} & Text & Pocket & Text+Pocket & SG & SG+Pocket & ST & ST+Pocket & ST+SG & ST+SG &ST+SG   \\
& &  &  &  &  & & & &+Text &+Pocket   \\
\midrule
DS1        & 1      & 1               & 0.9568          & 1      & 1               & 1      & 1               & 1      & 1      & 1      \\
DS1.2      & 1      & 1               & 1               & 1      & 1               & 1      & 1               & 1      & 1      & 1      \\
DS2        & 1      & 1               & 0.9305          & 1      & 1               & 1      & 1               & 1      & 1      & 1      \\
DS3        & 0.7678 & \textbf{0.9702} & 0.8519          & 0.8099 & 0.9234 & 0.8648 & 0.9344 & 0.8284 & 0.8538 & 0.9051 \\
DS4        & 0.7522 & \textbf{0.9590}  & 0.8238          & 0.8061 & 0.9102 & 0.8408 & 0.9179 & 0.801  & 0.8293 & 0.8907 \\
DS5 \& 5.2 & \textbf{0.6511} & 0.5749          & 0.6436          & 0.5434 & 0.5535          & 0.5608 & 0.5741          & 0.5485 & 0.6375 & 0.5675 \\
DS6        & 0.5605 & 0.5828          & 0.5806          & 0.46   & 0.5684          & 0.6142 & \textbf{0.6187}          & 0.5935 & 0.6108 & 0.6139 \\
DS6.2      & 0.5602 & 0.5828          & 0.5805          & 0.4598 & 0.5683          & 0.6141 & \textbf{0.6187}          & 0.5933 & 0.6107 & 0.6139 \\
DS7        & 0.8469 & 0.8193          & \textbf{0.8487} & 0.7552 & 0.7967          & 0.7723 & 0.8011          & 0.7446 & 0.8398 & 0.7928\\
\bottomrule
\end{tabular}
}
}
\end{table}

\begin{table*}[!htbp]
\centering
\caption{Receiver Operating Characteristic Area Under the Curve (AUC) Scores for the ProSPECCTs datasets using alternative to sequence embeddings from OneProt-5. For the last two columns concatenated embeddings are used for prediction.}
\label{s-tab:prospecct_auc_emb}
\small  % Reduced font size for better fit
\begin{tabular}{l*{8}{c}}  % Changed to center alignment for numbers
\toprule
\multirow{2}{*}{\textbf{Dataset}} 
 & Structure & Pocket & Sequence \& & Sequence, \\
 & Only & Only & Structure & Structure \& Pocket \\
\midrule
DS1  & 1.000 & 1.000 & 1.000 & 1.000 \\
DS1.2  & 1.000 & 1.000 & 1.000 & 1.000 \\
DS2   & 1.000 & 0.998 & 0.998 & 1.000 \\
DS3  & \textbf{0.531} & 0.495 & 0.495 & 0.505 \\
DS4  & \textbf{0.692} & 0.620 & 0.620 & 0.673 \\
DS5   & 0.603 & 0.598 & \textbf{0.599} & 0.613 \\
DS6  & 0.570 & 0.566 & 0.567 & 0.565 \\
DS6.2  & 0.573 & 0.577 & \textbf{0.578} & 0.571 \\
DS7  & 0.795 & \textbf{0.856} & \textbf{0.856} & 0.839 \\
\bottomrule
\end{tabular}
\end{table*}
\clearpage
\begin{table}[H]
\captionsetup{skip=1pt}
\centering
\caption{Numerical values corresponding to boxplots in Figures \ref{fig:topenzyme_boxplot} and \ref{fig:topenzyme_boxplot}. Q1 and Q3 correspond to the first and third quartiles of the data respectively. Column outliers correspond to the count of values below Q1-1.5$\times$IQR.  ST corresponds to Structure Token modality, SG corresponds to Structure Graph modality, '+' indicates a combination of multiple modalities.}\label{s-tab:topenzyme_outliers}
\makebox[\textwidth][c]{ 
\resizebox{1.0\linewidth}{!}{
\begin{tabular}{lcccccc}
\toprule
Model        & Q1   & Median & Q3   & IQR  & Outliers & Total Points \\
\midrule
TopEC                     & 0.50 & 0.75 & 0.97 & 0.47 & 0  & 826 \\
CLEAN                     & 0.57 & 0.95 & 1.00 & 0.43 & 0  & 826 \\
ESM2                      & 0.67 & 0.94 & 1.00 & 0.33 & 50 & 826 \\
ESM-IF                    & 0.62 & 0.92 & 1.00 & 0.38 & 55 & 826 \\
OpenFold                  & 0.69 & 0.95 & 1.00 & 0.31 & 54 & 826 \\
ProTrek-35M               & 0.61 & 0.89 & 0.99 & 0.38 & 52 & 826 \\
ProTrek-650M              & 0.75 & 0.95 & 1.00 & 0.25 & 58 & 826 \\
ST+SG+Pocket+Text (OneProt-5)                 & 0.71 & 0.97 & 1.00 & 0.29 & 51 & 826 \\
SG+Pocket+Text (OneProt-4)                 & 0.70 & 0.96 & 1.00 & 0.30 & 53 & 826 \\
Text Only                 & 0.70 & 0.94 & 1.00 & 0.30 & 60 & 826 \\
Pocket+Text               & 0.70 & 0.96 & 1.00 & 0.30 & 58 & 826 \\
SG+Text                   & 0.68 & 0.95 & 1.00 & 0.32 & 53 & 826 \\
ST+Text                   & 0.70 & 0.96 & 1.00 & 0.30 & 62 & 826 \\
ST+SG+Text                & 0.72 & 0.95 & 1.00 & 0.28 & 46 & 826 \\
ST+SG+Pocket              & 0.70 & 0.94 & 1.00 & 0.30 & 51 & 826 \\
ST+Pocket+Text  & 0.73 & 0.96 & 1.00 & 0.27 & 55 & 826 \\
ST+Pocket                 & 0.71 & 0.96 & 1.00 & 0.29 & 52 & 826 \\
ST Only                   & 0.68 & 0.93 & 1.00 & 0.32 & 68 & 826 \\
ST+SG                     & 0.62 & 0.93 & 1.00 & 0.38 & 67 & 826 \\
Pocket Only               & 0.65 & 0.92 & 1.00 & 0.35 & 55 & 826 \\
SG+Pocket                 & 0.61 & 0.91 & 1.00 & 0.39 & 62 & 826 \\
\bottomrule
\end{tabular}
}}\
\end{table}

\begin{figure}[H]
    \centering
\makebox[\textwidth][c]{ 
    \includegraphics[width=1.35\linewidth]{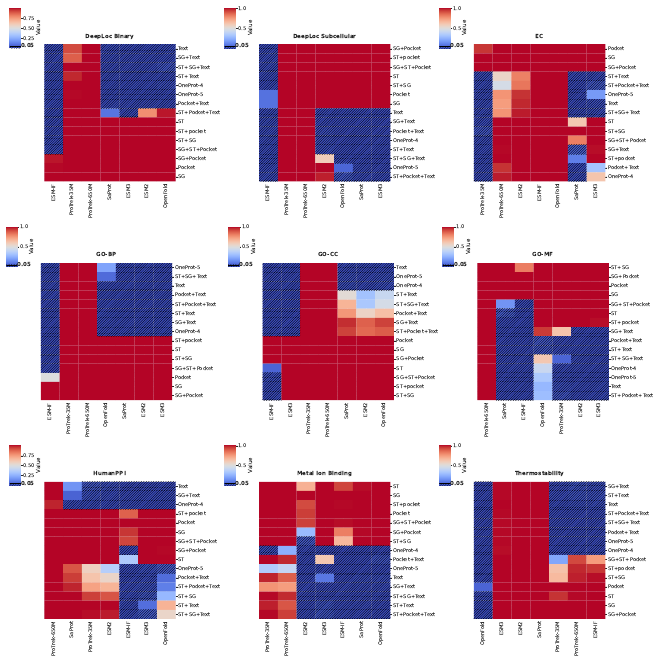}
    }

    \caption{\textbf {Heatmaps of $p$-values according to one-sided Wilcoxon rank-sum test.} The alternative hypothesis of OneProt models (vertical axis) outperforming baseline models (horizontal axis) according to metric values from Tables \ref{tab: saprot tasks},\ref{tab: saprot tasks 1}, where for binary classification accuracy was compared. Striped pattern stands for the values $p<0.05$, when the null hypothesis of OneProt being non-superior was rejected. }\label{s-fig:heatmaps_p}
\end{figure}

\begin{figure}[H]
    \centering
\makebox[\textwidth][c]{ 
    \includegraphics[width=1.3\linewidth]{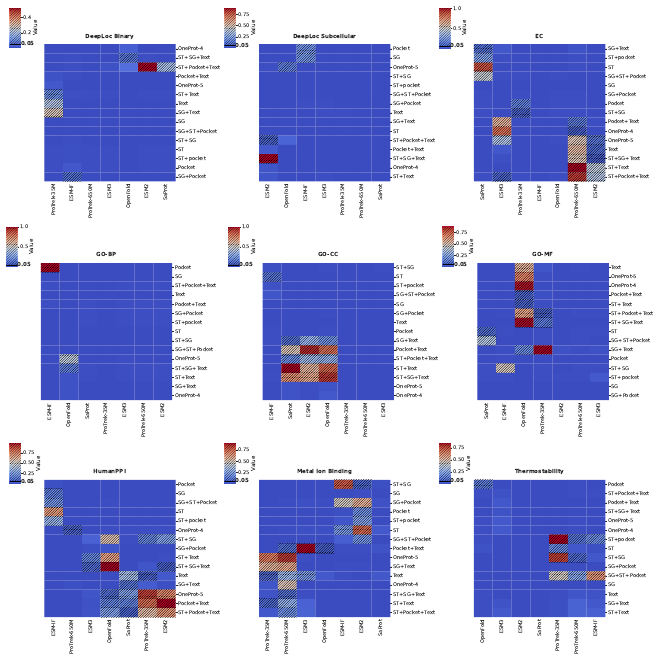}
    }

    \caption{\textbf{Heatmaps of $p$-values according to two-sided Wilcoxon rank-sum test.} The alternative hypothesis of OneProt models (vertical axis) performing differently from baseline models (horizontal axis) according to metric values from Tables \ref{tab: saprot tasks},\ref{tab: saprot tasks 1}, where for binary classification accuracy was compared. Striped pattern stands for the values $p\geq 0.05$, when the null hypothesis of OneProt being the same as the baseline was not rejected.}\label{s-fig:heatmaps_2p}
\end{figure}

\begin{figure}[H]
    \centering
\makebox[\textwidth][c]{ 
    \includegraphics[width=1.2\linewidth]{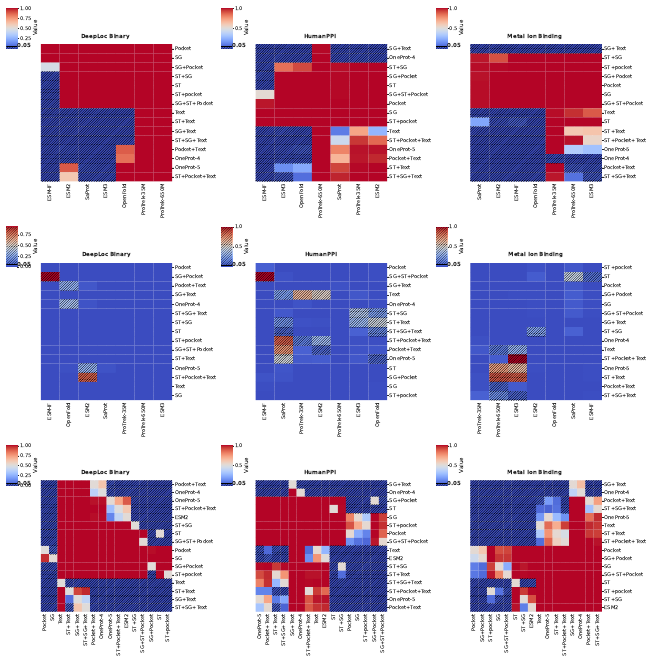}
    }

    \caption{\textbf{Heatmaps of $p$-values for Area Under Receiver Operating Characteristic curve metrics.} One-sided Wilcoxon rank-sum test with the alternative hypothesis of OneProt (vertical axis) outperforming baseline models (horizontal axis), striped pattern corresponding to values $p<0.05$  (upper panel). Two-sided Wilcoxon rank-sum test with the alternative hypothesis of OneProt performing differently than baseline models, striped pattern corresponding to values $p\geq 0.05$  (middle panel). One-sided Wilcoxon rank-sum test for OneProt ablations with the alternative hypothesis of the models on the vertical axis outperforming the models on the horizontal axis (bottom panel). Striped pattern stands for the values $p<0.05$.}\label{s-fig:heatmaps_AUC}
\end{figure}

\begin{figure}[H]
    \centering
\makebox[\textwidth][c]{ 
    \includegraphics[width=1.3\linewidth]{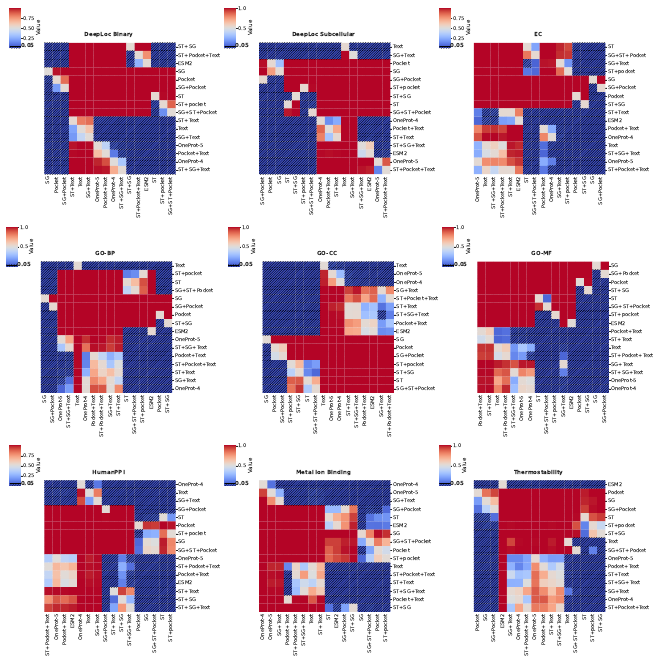}
    }
    \caption{\textbf {Heatmaps of $p$-values according to one-sided Wilcoxon rank-sum test for OneProt ablations.} The alternative hypothesis of the models on the vertical axis outperforming the models on the horizontal according to metric values from Tables \ref{tab: saprot tasks},\ref{tab: saprot tasks 1}, where for binary classification accuracy was compared. Striped pattern stands for the values $p<0.05$, when the null hypothesis of model on the vertical axis being non-superior was rejected.}\label{s-fig:heatmaps_ablations}
\end{figure}

%\subsection{Enzyme function prediction}\label{supp:topenzyme}
\begin{figure}[H]
    \centering
\makebox[\textwidth][c]{ 
    \includegraphics[width=1.3\linewidth]{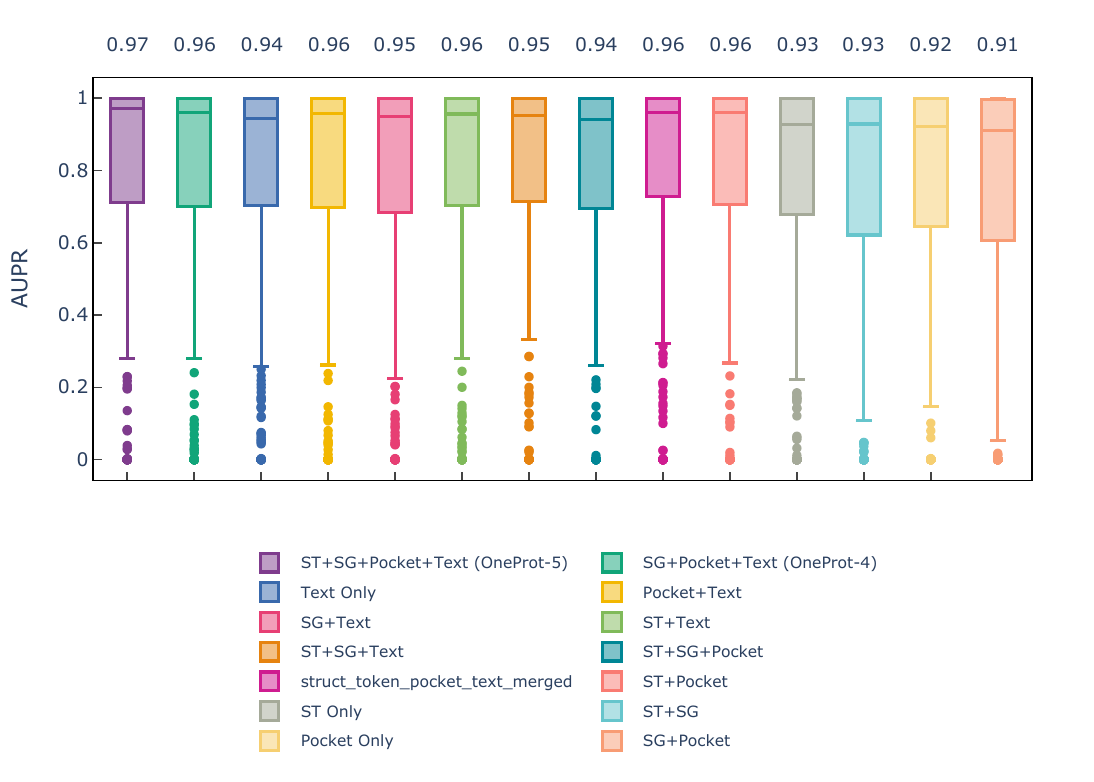}
    }
    \caption{Performance comparison of OneProt ablations on the TopEnzyme dataset using boxplots of the Area Under Precision Recall curve (AUPR) distributions.  ST corresponds to Structure Token modality, SG corresponds to Structure Graph modality, '+' indicates a combination of multiple modalities.
    %{\bf Alt Text:} Box plots displaying the AUPR distributions of the TopEnzyme EC numbers prediction across 6 methods with median AUPR value stated above each plot.
  }
    \label{s-fig:topenzyme_boxplot}
\end{figure}
%\subsection{Biological Contrastive Learning}\label{supp:bio}

\begin{figure}[H]
    \centering
    %\hspace*{-0.5\linewidth}
    \makebox[\textwidth][c]{ 
    \includegraphics[width=1.6\linewidth]{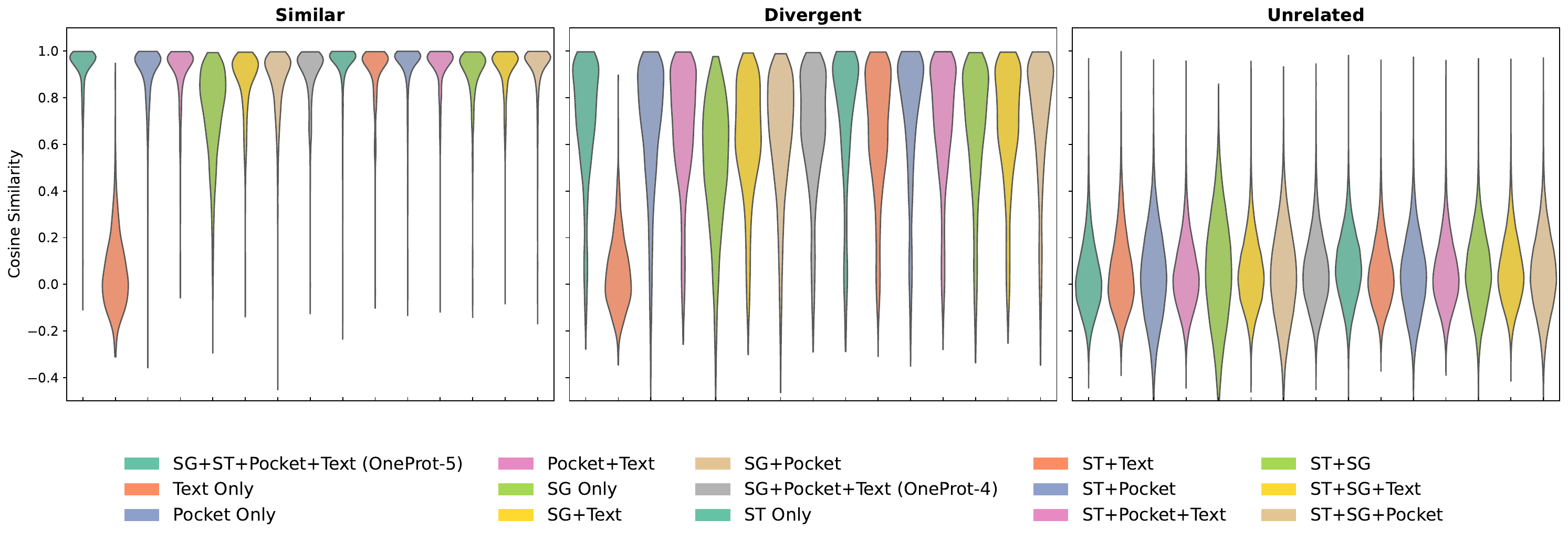}
    }
    \caption{Cosine Similarity distributions for OneProt ablations. The plot shows the similarity of a given protein to three groups: the 50 most evolutionarily similar proteins, the 50 most evolutionarily divergent sequences, and 1000 unrelated sequences.ST corresponds to Structure Token modality, SG corresponds to Structure Graph modality, '+' indicates a combination of multiple modalities.}
    \label{s-fig:similarity}
\end{figure}
%\subsection{ProSPECCTs}\label{supp:prosspects}

%\clearpage
\begin{table}[H]\caption{Table of ranges (Min, Max), 0.25 (Q1), 0.5 (Median), 0.75 (Q3), Inter Quantile Range (IQR = Q3 - Q1) for metrics of different models on DeepLoc2, DeepLoc10 (accuracy), EC (Fmax) tasks. Task and modality names as in Table \ref{s-tab:downstream}.}\label{s-fig:iqr1}
\resizebox{0.56\linewidth}{!}{
\begin{tabular}{|l|llllll|}
\hline
\textbf{DeepLoc2}        &       &         &        &         &       &         \\
\textbf{Accuracy} & \textbf{Min}   & \textbf{Q1}      & \textbf{Median} & \textbf{Q3}      & \textbf{Max}   & \textbf{IQR}     \\
\hline
OnepProt-5      & 0.92  & 0.923 & 0.924  & 0.925   & 0.925 & 0.002 \\
Text Only       & 0.926 & 0.928   & 0.928  & 0.931   & 0.933 & 0.003   \\
Pocket Only     & 0.826 & 0.831   & 0.837  & 0.840    & 0.842 & 0.009   \\
Pocket+Text     & 0.92  & 0.922 & 0.923 & 0.925   & 0.926 & 0.003 \\
SG only         & 0.813 & 0.825   & 0.830 & 0.834   & 0.835 & 0.009   \\
SG+Text         & 0.923 & 0.927   & 0.929  & 0.932   & 0.934 & 0.005   \\
SG+Pocket       & 0.833 & 0.835   & 0.837  & 0.846   & 0.848 & 0.011   \\
OneProt-4       & 0.916 & 0.920  & 0.921  & 0.924   & 0.928 & 0.004  \\
ST only         & 0.864 & 0.865   & 0.867 & 0.868   & 0.876 & 0.003   \\
ST+Text         & 0.923 & 0.925   & 0.929 & 0.929   & 0.931 & 0.004   \\
ST+Pocket       & 0.868 & 0.873   & 0.876  & 0.88    & 0.881 & 0.007   \\
ST+Pocket+Text  & 0.908 & 0.912   & 0.913  & 0.916   & 0.916 & 0.004   \\
ST+SG           & 0.885 & 0.888   & 0.891  & 0.895   & 0.897 & 0.007   \\
ST+SG+Text      & 0.917 & 0.920    & 0.921  & 0.921   & 0.924 & 0.001   \\
ST+SG+Pocket    & 0.872 & 0.874   & 0.880   & 0.883   & 0.885 & 0.009   \\
ProTrek-35M     & 0.924 & 0.929 & 0.933 & 0.933 & 0.935 & 0.004   \\
ProTrek-650M    & 0.95  & 0.951 & 0.953 & 0.954 & 0.955 & 0.003  \\
ESM-2           & 0.906 & 0.911 & 0.915  & 0.916 & 0.918 & 0.005   \\
SaProt          & 0.911 & 0.907   & 0.910   & 0.913   & 0.913 & 0.006   \\
ESM-3           & 0.905 & 0.906   & 0.906  & 0.908   & 0.91  & 0.002   \\
ESM-IF          & 0.838 & 0.844   & 0.848 & 0.848   & 0.854 & 0.004   \\
OpenFold        & 0.912 & 0.914   & 0.918  & 0.92    & 0.921 & 0.006   \\
OneProt-4 matched &0.913 &0.914 &0.918 &0.922 &0.926 & 0.008\\
\hline
\textbf{DeepLoc10}       &       &         &        &         &       &         \\
\textbf{Accuracy}                & \textbf{Min}   & \textbf{Q1}      & \textbf{Median} & \textbf{Q3}      & \textbf{Max}   & \textbf{IQR}     \\
\hline
OnepProt-5      & 0.799 & 0.801   & 0.802  & 0.805   & 0.806 & 0.004   \\
Text Only       & 0.823 & 0.828   & 0.829  & 0.834   & 0.834 & 0.006   \\
Pocket Only     & 0.604 & 0.617   & 0.6245 & 0.628   & 0.629 & 0.011   \\
Pocket+Text     & 0.809 & 0.812   & 0.816  & 0.819   & 0.821 & 0.007   \\
SG only         & 0.608 & 0.615   & 0.620   & 0.627   & 0.628 & 0.012   \\
SG+Text         & 0.817 & 0.820    & 0.823 & 0.824   & 0.828 & 0.004   \\
SG+Pocket       & 0.623 & 0.629 & 0.631 & 0.631 & 0.632 & 0.003  \\
OneProt-4       & 0.812 & 0.813   & 0.819  & 0.820    & 0.821 & 0.007   \\
ST only         & 0.658 & 0.665 & 0.670 & 0.671 & 0.680  & 0.006  \\
ST+Text         & 0.809 & 0.814   & 0.818 & 0.818   & 0.819 & 0.004   \\
ST+Pocket       & 0.64  & 0.649   & 0.653 & 0.657   & 0.659 & 0.008   \\
ST+Pocket+Text  & 0.798 & 0.802   & 0.805  & 0.808   & 0.812 & 0.006   \\
ST+SG           & 0.67  & 0.675 & 0.683 & 0.686 & 0.69  & 0.011  \\
ST+SG+Text      & 0.803 & 0.807   & 0.810 & 0.813   & 0.814 & 0.006   \\
ST+SG+Pocket    & 0.656 & 0.658   & 0.662 & 0.664   & 0.669 & 0.006   \\
ProTrek-35M     & 0.834 & 0.835   & 0.836  & 0.839   & 0.844 & 0.004   \\
ProTrek-650M    & 0.902 & 0.908 & 0.911 & 0.911   & 0.913 & 0.003 \\
ESM-2           & 0.803 & 0.807   & 0.811  & 0.813   & 0.813 & 0.006   \\
SaProt          & 0.783 & 0.788   & 0.791  & 0.795   & 0.795 & 0.007   \\
ESM-3           & 0.756 & 0.759   & 0.763  & 0.765   & 0.77  & 0.006   \\
ESM-IF          & 0.609 & 0.609   & 0.613  & 0.618   & 0.627 & 0.009   \\
OpenFold        & 0.791 & 0.798   & 0.800    & 0.802   & 0.805 & 0.004   \\
OneProt-4 matched &0.800 &0.807 &0.809 &0.811 &0.812 & 0.004\\
\hline
\textbf{EC}              &       &         &        &         &       &         \\
\textbf{Fmax}                & \textbf{Min}   & \textbf{Q1}      & \textbf{Median} & \textbf{Q3}      & \textbf{Max}   & \textbf{IQR}     \\
\hline
OnepProt-5      & 0.869 & 0.870    & 0.872  & 0.879   & 0.884 & 0.009   \\
Text Only       & 0.872 & 0.874   & 0.875  & 0.876   & 0.882 & 0.002   \\
Pocket Only     & 0.839 & 0.841   & 0.843  & 0.844   & 0.849 & 0.003   \\
Pocket+Text     & 0.867 & 0.868   & 0.873  & 0.875   & 0.876 & 0.007   \\
SG only         & 0.691 & 0.696   & 0.700    & 0.701   & 0.703 & 0.005   \\
SG+Text         & 0.86  & 0.864   & 0.868  & 0.869   & 0.871 & 0.005   \\
SG+Pocket       & 0.826 & 0.829   & 0.829  & 0.83    & 0.83  & 0.001   \\
OneProt-4       & 0.867 & 0.8685  & 0.871  & 0.873   & 0.876 & 0.0045  \\
ST only         & 0.859 & 0.861   & 0.862  & 0.865   & 0.868 & 0.004   \\
ST+Text         & 0.871 & 0.876   & 0.878  & 0.878   & 0.882 & 0.002   \\
ST+Pocket       & 0.861 & 0.863 & 0.866  & 0.867 & 0.868 & 0.004  \\
ST+Pocket+Text  & 0.871 & 0.872   & 0.8765 & 0.879   & 0.881 & 0.007   \\
ST+SG           & 0.845 & 0.847   & 0.851  & 0.852   & 0.853 & 0.005   \\
ST+SG+Text      & 0.873 & 0.874   & 0.875  & 0.878   & 0.878 & 0.004   \\
ST+SG+Pocket    & 0.855 & 0.858   & 0.861 & 0.865   & 0.869 & 0.007   \\
ProTrek-35M     & 0.841 & 0.845   & 0.846  & 0.847   & 0.850  & 0.002   \\
ProTrek-650M    & 0.867 & 0.873   & 0.878 & 0.879   & 0.880  & 0.006   \\
ESM-2           & 0.874 & 0.876   & 0.878  & 0.879   & 0.885 & 0.003   \\
SaProt          & 0.856 & 0.862   & 0.863  & 0.866   & 0.867 & 0.004   \\
ESM-3           & 0.865 & 0.870    & 0.871  & 0.873   & 0.877 & 0.003   \\
ESM-IF          & 0.886 & 0.895  & 0.897 & 0.899 & 0.905 & 0.004 \\
OpenFold        & 0.882 & 0.885   & 0.889 & 0.889   & 0.892 & 0.004   \\
OneProt-4 matched & 0.862 & 0.864 & 0.869 & 0.870 & 0.872 & 0.006\\
\hline
\end{tabular}
}
\end{table}

\begin{table}[H]\caption{Table of ranges (Min, Max), 0.25 (Q1), 0.5 (Median), 0.75 (Q3), Inter Quantile Range (IQR = Q3 - Q1) for metrics of different models on GO-BP, GO-CC, GO-MF (Fmax) tasks.Task and modality names as in Table \ref{s-tab:downstream}.}\label{s-fig:iqr2}
\resizebox{0.55\linewidth}{!}{
\begin{tabular}{|l|llllll|}
\hline
\textbf{GO-BP}           &       &         &        &         &       &         \\
\textbf{Fmax}                & \textbf{Min}   & \textbf{Q1}      & \textbf{Median} & \textbf{Q3}      & \textbf{Max}   & \textbf{IQR}     \\
\hline
OnepProt-5      & 0.486 & 0.491   & 0.492  & 0.493   & 0.496 & 0.002   \\
Text Only       & 0.499 & 0.502   & 0.503  & 0.504   & 0.505 & 0.002   \\
Pocket Only     & 0.433 & 0.435   & 0.437  & 0.439   & 0.441 & 0.004   \\
Pocket+Text     & 0.494 & 0.494   & 0.497 & 0.499   & 0.500   & 0.005   \\
SG only         & 0.362 & 0.364   & 0.367  & 0.369   & 0.370  & 0.005   \\
SG+Text         & 0.492 & 0.492   & 0.495  & 0.499   & 0.499 & 0.007   \\
SG+Pocket       & 0.419 & 0.425   & 0.426 & 0.426   & 0.428 & 0.001   \\
OneProt-4       & 0.491 & 0.493 & 0.494 & 0.496 & 0.500   & 0.003  \\
ST only         & 0.456 & 0.456   & 0.458  & 0.460   & 0.466 & 0.004   \\
ST+Text         & 0.491 & 0.495   & 0.496  & 0.498   & 0.498 & 0.003   \\
ST+Pocket       & 0.458 & 0.458   & 0.460   & 0.462   & 0.466 & 0.004   \\
ST+Pocket+Text  & 0.493 & 0.494   & 0.496  & 0.498   & 0.503 & 0.004   \\
ST+SG           & 0.446 & 0.447   & 0.449  & 0.449   & 0.451 & 0.002   \\
ST+SG+Text      & 0.491 & 0.491   & 0.493  & 0.494   & 0.495 & 0.003   \\
ST+SG+Pocket    & 0.457 & 0.457   & 0.458  & 0.461   & 0.466 & 0.004   \\
ProTrek-35M     & 0.504 & 0.513  & 0.515 & 0.518 & 0.519 & 0.005 \\
ProTrek-650M    & 0.53  & 0.537   & 0.538  & 0.541   & 0.543 & 0.004   \\
ESM-2           & 0.474 & 0.476   & 0.479  & 0.481   & 0.483 & 0.005   \\
SaProt          & 0.469 & 0.472   & 0.473  & 0.474   & 0.477 & 0.002   \\
ESM-3           & 0.478 & 0.48    & 0.481  & 0.483   & 0.488 & 0.003   \\
ESM-IF          & 0.432 & 0.432   & 0.438  & 0.439   & 0.44  & 0.007   \\
OpenFold        & 0.487 & 0.491 & 0.492 & 0.492   & 0.493 & 0.001 \\
OneProt-4 matched &0.488 &0.488 &0.491 & 0.494 &0.494 & 0.006\\
\hline
\textbf{GO-CC}           &       &         &        &         &       &         \\
\textbf{Fmax}                & \textbf{Min}   & \textbf{Q1}      & \textbf{Median} & \textbf{Q3}      & \textbf{Max}   & \textbf{IQR}     \\
\hline
OnepProt-5      & 0.551 & 0.552   & 0.556  & 0.56    & 0.562 & 0.008   \\
Text Only       & 0.553 & 0.556 & 0.563 & 0.564   & 0.567 & 0.008 \\
Pocket Only     & 0.471 & 0.471   & 0.472  & 0.475   & 0.478 & 0.004   \\
Pocket+Text     & 0.537 & 0.545   & 0.545 & 0.551   & 0.554 & 0.006   \\
SG only         & 0.428 & 0.442   & 0.448 & 0.454  & 0.462 & 0.0115  \\
SG+Text         & 0.541 & 0.543   & 0.5435 & 0.545   & 0.552 & 0.002   \\
SG+Pocket       & 0.469 & 0.471   & 0.4735 & 0.476   & 0.477 & 0.005   \\
OneProt-4       & 0.544 & 0.552   & 0.555  & 0.558   & 0.565 & 0.006   \\
ST only         & 0.482 & 0.489   & 0.495  & 0.504   & 0.517 & 0.015   \\
ST+Text         & 0.54  & 0.545 & 0.548 & 0.554 & 0.559 & 0.009   \\
ST+Pocket       & 0.485 & 0.498   & 0.499  & 0.507   & 0.518 & 0.009   \\
ST+Pocket+Text  & 0.541 & 0.541   & 0.546  & 0.549   & 0.55  & 0.008   \\
ST+SG           & 0.495 & 0.496   & 0.502  & 0.512   & 0.513 & 0.016   \\
ST+SG+Text      & 0.545 & 0.547   & 0.548  & 0.549   & 0.554 & 0.002   \\
ST+SG+Pocket    & 0.491 & 0.493   & 0.494  & 0.504   & 0.512 & 0.011   \\
ProTrek-35M     & 0.568 & 0.58    & 0.583  & 0.589   & 0.594 & 0.009   \\
ProTrek-650M    & 0.611 & 0.611   & 0.618  & 0.621   & 0.623 & 0.01    \\
ESM-2           & 0.543 & 0.548   & 0.551  & 0.552   & 0.556 & 0.004   \\
SaProt          & 0.54  & 0.548 & 0.549 & 0.550    & 0.554 & 0.002 \\
ESM-3           & 0.508 & 0.529   & 0.531  & 0.536   & 0.541 & 0.007   \\
ESM-IF          & 0.479 & 0.485   & 0.489 & 0.491   & 0.496 & 0.006   \\
OpenFold        & 0.541 & 0.546   & 0.548  & 0.551   & 0.553 & 0.005   \\
OneProt-4 matched &0.534 &0.544 &0.548 &0.557 &0.559 & 0.013\\
\hline
\textbf{GO-MF}           &       &         &        &         &       &         \\
\textbf{Fmax}                & \textbf{Min}   & \textbf{Q1}      & \textbf{Median} & \textbf{Q3}      & \textbf{Max}   & \textbf{IQR}     \\
\hline
OnepProt-5      & 0.653 & 0.654   & 0.656  & 0.657   & 0.659 & 0.003   \\
Text Only       & 0.652 & 0.653   & 0.659  & 0.660    & 0.660  & 0.007   \\
Pocket Only     & 0.600   & 0.604   & 0.605 & 0.608   & 0.608 & 0.004   \\
Pocket+Text     & 0.652 & 0.659   & 0.660   & 0.661   & 0.662 & 0.002   \\
SG only         & 0.463 & 0.464   & 0.467  & 0.472   & 0.474 & 0.008   \\
SG+Text         & 0.647 & 0.647   & 0.651 & 0.653   & 0.659 & 0.006   \\
SG+Pocket       & 0.583 & 0.584   & 0.590   & 0.593   & 0.596 & 0.009   \\
OneProt-4       & 0.654 & 0.655   & 0.656  & 0.656   & 0.656 & 0.001   \\
ST only         & 0.622 & 0.628   & 0.632 & 0.632   & 0.637 & 0.004   \\
ST+Text         & 0.657 & 0.659   & 0.661 & 0.663   & 0.665 & 0.004   \\
ST+Pocket       & 0.632 & 0.636   & 0.637  & 0.638   & 0.643 & 0.002   \\
ST+Pocket+Text  & 0.648 & 0.656   & 0.657  & 0.657   & 0.664 & 0.001   \\
ST+SG           & 0.607 & 0.607   & 0.608 & 0.612   & 0.613 & 0.005   \\
ST+SG+Text      & 0.647 & 0.652   & 0.655  & 0.657   & 0.658 & 0.005   \\
ST+SG+Pocket    & 0.623 & 0.624   & 0.627 & 0.628   & 0.631 & 0.004   \\
ProTrek-35M     & 0.648 & 0.649   & 0.651  & 0.653   & 0.653 & 0.004   \\
ProTrek-650M    & 0.669 & 0.669   & 0.676  & 0.678   & 0.684 & 0.009   \\
ESM-2           & 0.64  & 0.643   & 0.645  & 0.647   & 0.649 & 0.004   \\
SaProt          & 0.619 & 0.620    & 0.624  & 0.628   & 0.635 & 0.008   \\
ESM-3           & 0.63  & 0.640  & 0.642  & 0.644 & 0.648 & 0.004 \\
ESM-IF          & 0.606 & 0.608   & 0.610   & 0.613   & 0.619 & 0.005   \\
OpenFold        & 0.649 & 0.652 & 0.655  & 0.657 & 0.664 & 0.005\\
OneProt-4 matched &0.648 &0.648 &0.652 &0.653 &0.656 & 0.005\\
\hline
\end{tabular}
}
\end{table}

\begin{table}[H]\caption{Table of ranges (Min, Max), 0.25 (Q1), 0.5 (Median), 0.75 (Q3), Inter Quantile Range (IQR = Q3 - Q1) for metrics of different models on HumanPPI, Metal Ion Binding (accuracy), ThermoStability (Spearman's $\rho$) tasks. Task and modality names as in Table \ref{s-tab:downstream}. }\label{s-fig:iqr3}
\resizebox{0.54\linewidth}{!}{
\begin{tabular}{|l|llllll|}
\hline
\textbf{HumanPPI}        &       &         &        &         &       &         \\
                & \textbf{Min}   & \textbf{Q1}      & \textbf{Median} & \textbf{Q3}      & \textbf{Max}   & \textbf{IQR}     \\
\hline
OnepProt-5      & 0.811 & 0.856   & 0.867  & 0.872   & 0.883 & 0.016   \\
Text Only       & 0.856 & 0.861   & 0.875  & 0.883   & 0.900  & 0.022   \\
Pocket Only     & 0.744 & 0.75    & 0.759 & 0.767   & 0.772 & 0.017   \\
Pocket+Text     & 0.833 & 0.856   & 0.861  & 0.867   & 0.894 & 0.011   \\
SG only         & 0.756 & 0.761   & 0.767  & 0.794   & 0.794 & 0.033   \\
SG+Text         & 0.833 & 0.856   & 0.872  & 0.883   & 0.889 & 0.027   \\
SG+Pocket       & 0.800   & 0.800     & 0.811  & 0.822   & 0.828 & 0.022   \\
OneProt-4       & 0.85  & 0.889   & 0.889  & 0.900     & 0.906 & 0.011   \\
ST only         & 0.767 & 0.783   & 0.794  & 0.806   & 0.817 & 0.023   \\
ST+Text         & 0.817 & 0.838  & 0.844  & 0.856   & 0.861 & 0.018  \\
ST+Pocket       & 0.75  & 0.771 & 0.772  & 0.794   & 0.833 & 0.023 \\
ST+Pocket+Text  & 0.833 & 0.844   & 0.861  & 0.867   & 0.883 & 0.023   \\
ST+SG           & 0.828 & 0.844   & 0.856  & 0.861   & 0.872 & 0.017   \\
ST+SG+Text      & 0.833 & 0.844   & 0.85   & 0.856   & 0.872 & 0.012   \\
ST+SG+Pocket    & 0.756 & 0.761   & 0.767 & 0.783   & 0.806 & 0.022   \\
ProTrek-35M     & 0.839 & 0.856   & 0.859 & 0.878   & 0.894 & 0.022   \\
ProTrek-650M    & 0.878 & 0.894   & 0.906  & 0.911   & 0.928 & 0.017   \\
ESM-2           & 0.844 & 0.855  & 0.861  & 0.868 & 0.872 & 0.013 \\
SaProt          & 0.850  & 0.861   & 0.870   & 0.883   & 0.900   & 0.022   \\
ESM-3           & 0.811 & 0.825 & 0.834 & 0.847   & 0.861 & 0.022 \\
ESM-IF          & 0.756 & 0.782 & 0.794  & 0.794   & 0.806 & 0.012 \\
OpenFold        & 0.806 & 0.817   & 0.85   & 0.861   & 0.867 & 0.044   \\
OneProt-4 matched &0.839 &0.855 &0.859 &0.863 &0.867 & 0.008\\
\hline
\textbf{MetalIonBinding} &       &         &        &         &       &         \\
                & \textbf{Min}   & \textbf{Q1}      & \textbf{Median} & \textbf{Q3}      & \textbf{Max}   & \textbf{IQR}     \\
\hline
OnepProt-5      & 0.732 & 0.753   & 0.767  & 0.770    & 0.782 & 0.017   \\
Text Only       & 0.709 & 0.719   & 0.754  & 0.759   & 0.762 & 0.040    \\
Pocket Only     & 0.642 & 0.650    & 0.66   & 0.675   & 0.677 & 0.025   \\
Pocket+Text     & 0.72  & 0.722   & 0.723 & 0.725   & 0.744 & 0.003   \\
SG only         & 0.615 & 0.630    & 0.640   & 0.645   & 0.662 & 0.015   \\
SG+Text         & 0.746 & 0.752   & 0.763  & 0.764   & 0.768 & 0.012   \\
SG+Pocket       & 0.656 & 0.674   & 0.682  & 0.683   & 0.716 & 0.009   \\
OneProt-4       & 0.765 & 0.770  & 0.774 & 0.776 & 0.780  & 0.006 \\
ST only         & 0.653 & 0.657   & 0.668  & 0.692   & 0.719 & 0.035   \\
ST+Text         & 0.731 & 0.741   & 0.747  & 0.756   & 0.762 & 0.015   \\
ST+Pocket       & 0.632 & 0.643  & 0.660 & 0.675 & 0.687 & 0.032 \\
ST+Pocket+Text  & 0.729 & 0.740    & 0.744  & 0.755   & 0.758 & 0.015   \\
ST+SG           & 0.675 & 0.678   & 0.683  & 0.695   & 0.696 & 0.017   \\
ST+SG+Text      & 0.743 & 0.744 & 0.747 & 0.748 & 0.75  & 0.004   \\
ST+SG+Pocket    & 0.626 & 0.638   & 0.642  & 0.678   & 0.701 & 0.040    \\
ProTrek-35M     & 0.746 & 0.749   & 0.765  & 0.770    & 0.771 & 0.021   \\
ProTrek-650M    & 0.692 & 0.749   & 0.765  & 0.779   & 0.783 & 0.030    \\
ESM-2           & 0.659 & 0.663   & 0.673  & 0.683   & 0.693 & 0.020    \\
SaProt          & 0.686 & 0.708 & 0.716  & 0.720    & 0.732 & 0.012 \\
ESM-3           & 0.696 & 0.707   & 0.737  & 0.741   & 0.744 & 0.034   \\
ESM-IF          & 0.672 & 0.680    & 0.689  & 0.696   & 0.711 & 0.016   \\
OpenFold        & 0.708 & 0.713   & 0.717 & 0.723   & 0.731 & 0.010    \\
OneProt-4 matched &0.713 &0.743 &0.744 &0.755 &0.761 & 0.012\\
\hline
\textbf{ThermoStability} &       &         &        &         &       &         \\
                & \textbf{Min}   & \textbf{Q1}      & \textbf{Median} & \textbf{Q3}      & \textbf{Max}   & \textbf{IQR}     \\
\hline
OnepProt-5      & 0.657 & 0.668   & 0.676  & 0.681   & 0.682 & 0.013   \\
Text Only       & 0.65  & 0.651   & 0.656  & 0.66    & 0.663 & 0.009   \\
Pocket Only     & 0.572 & 0.573   & 0.601  & 0.619   & 0.628 & 0.046   \\
Pocket+Text     & 0.658 & 0.667 & 0.673 & 0.675   & 0.679 & 0.008 \\
SG only         & 0.612 & 0.614 & 0.616 & 0.618   & 0.623 & 0.004 \\
SG+Text         & 0.640      & 0.657        & 0.666       & 0.674        &  0.681     & 0.007       \\
SG+Pocket       & 0.64  & 0.657   & 0.666  & 0.674   & 0.681 & 0.017   \\
OneProt-4       & 0.656 & 0.668   & 0.670   & 0.672   & 0.673 & 0.004   \\
ST only         & 0.589 & 0.619   & 0.6285 & 0.636   & 0.639 & 0.017   \\
ST+Text         & 0.638 & 0.666   & 0.6715 & 0.674   & 0.677 & 0.008   \\
ST+Pocket       & 0.604 & 0.630    & 0.636  & 0.644   & 0.648 & 0.014   \\
ST+Pocket+Text  & 0.665 & 0.666   & 0.669  & 0.671   & 0.683 & 0.005   \\
ST+SG           & 0.628 & 0.630    & 0.635 & 0.644   & 0.648 & 0.014   \\
ST+SG+Text      & 0.66  & 0.665 & 0.672 & 0.674   & 0.675 & 0.009 \\
ST+SG+Pocket    & 0.632 & 0.638   & 0.644  & 0.646   & 0.646 & 0.008   \\
ProTrek-35M     & 0.623 & 0.629   & 0.637  & 0.647   & 0.653 & 0.018   \\
ProTrek-650M    & 0.628 & 0.643   & 0.649  & 0.65    & 0.655 & 0.007   \\
ESM-2           & 0.690  & 0.692   & 0.696  & 0.700     & 0.703 & 0.008   \\
SaProt          & 0.693 & 0.699   & 0.701  & 0.705   & 0.71  & 0.006   \\
ESM-3           & 0.669 & 0.686   & 0.688 & 0.702   & 0.712 & 0.016   \\
ESM-IF          & 0.637 & 0.638   & 0.645  & 0.650    & 0.653 & 0.012   \\
OpenFold        & 0.567 & 0.571   & 0.580   & 0.590    & 0.601 & 0.019  \\
OneProt-4 matched &0.635 &0.642 &0.645 &0.650 &0.668 & 0.008\\
\hline
\end{tabular}
}
\end{table}

% Please add the following required packages to your document preamble:
% \usepackage[table,xcdraw]{xcolor}
% Beamer presentation requires \usepackage{colortbl} instead of \usepackage[table,xcdraw]{xcolor}
% Please add the following required packages to your document preamble:
% \usepackage[table,xcdraw]{xcolor}
% Beamer presentation requires \usepackage{colortbl} instead of \usepackage[table,xcdraw]{xcolor}
\begin{table}[H]\caption{ Table of ranges (Min, Max), 0.25 (Q1), 0.5 (Median), 0.75 (Q3), Inter Quantile Range (IQR = Q3 - Q1) for AUC of different models on binary classification tasks: DeepLoc2, HumanPPI, Metal IonBinding) tasks. Task and modality names as in Table \ref{s-tab:downstream}.}\label{s-fig:iqr4}
\resizebox{0.6\linewidth}{!}{
% Please add the following required packages to your document preamble:
% \usepackage[table,xcdraw]{xcolor}
% Beamer presentation requires \usepackage{colortbl} instead of \usepackage[table,xcdraw]{xcolor}
\begin{tabular}{|l|llllll|}
\hline
\textbf{DeepLoc2 }          &                               &                               &                               &                               &                               &         \\
\textbf{AUC}                      & \textbf{Min}                           & \textbf{Q1}                            & \textbf{Median}                        & \textbf{Q3}                            & \textbf{Max}                           & \textbf{IQR}    \\
\hline
OnepProt-5            & 0.960                         & 0.962                         & 0.964                         & 0.964                         & 0.967                         & 0.002 \\
Text Only             & 0.971                         & 0.972                         & 0.973                         & 0.974                         & 0.975                         & 0.002 \\
Pocket Only           & 0.894                         & 0.898                         & 0.900                         & 0.902                         & 0.903                         & 0.005 \\
Pocket+Text           & 0.961                         & 0.963                         & 0.964                         & 0.966                         & 0.968                         & 0.003 \\
SG only               & 0.889                         & 0.893                         & 0.896                         & 0.900                         & 0.903                         & 0.008 \\
SG+Text               & 0.964                         & 0.967                         & 0.970                         & 0.972                         & 0.973                         & 0.005 \\
SG+Pocket             & 0.898                         & 0.901                         & 0.906                         & 0.907                         & 0.909                         & 0.006 \\
OneProt-4             & 0.960                         & 0.964                         & 0.965                         & 0.967                         & 0.969                         & 0.003 \\
ST only               & 0.924                         & 0.926                         & 0.929                         & 0.929                         & 0.930                         & 0.003 \\
ST+Text               & 0.967                         & 0.968                         & 0.969                         & 0.969                         & 0.970                         & 0.001 \\
ST+Pocket             & 0.916                         & 0.920                         & 0.922                         & 0.922                         & 0.928                         & 0.002 \\
ST+Pocket+Text        & 0.960                         & 0.961                         & 0.963                         & 0.965                         & 0.966                         & 0.004 \\
ST+SG                 & 0.940                         & 0.941                         & 0.943                         & 0.944                         & 0.944                         & 0.003 \\
ST+SG+Text            & 0.964                         & 0.967                         & 0.970                         & 0.971                         & 0.971                         & 0.004 \\
ST+SG+Pocket          & 0.928                         & 0.930                         & 0.930                         & 0.932                         & 0.934                         & 0.002 \\
ProTrek-35M           & 0.975                         & 0.977                         & 0.978                         & 0.979                         & 0.981                         & 0.002 \\
ProTrek-650M          & 0.986                         & 0.988                         & 0.988                         & 0.989                         & 0.990                         & 0.001 \\
ESM-2                 & 0.959                         & 0.963                         & 0.964                         & 0.965                         & 0.968                         & 0.002 \\
SaProt                & 0.913                         & 0.958                         & 0.959                         & 0.961                         & 0.963                         & 0.003 \\
ESM-3                 & 0.955                         & 0.956                         & 0.957                         & 0.958                         & 0.960                         & 0.002 \\
ESM-IF                & 0.902                         & 0.904                         & 0.904                         & 0.906                         & 0.909                         & 0.002 \\
OpenFold              & 0.962                         & 0.965                         & 0.966                         & 0.968                         & 0.968                         & 0.003 \\
\hline
\textbf{HumanPPI }          &                               &                               &                               &                               &                               &       \\
\textbf{AUC }                     & \textbf{Min}                           & \textbf{Q1}                            & \textbf{Median}                        & \textbf{Q3}                            & \textbf{Max}                           & \textbf{IQR}   \\
\hline
OnepProt-5            & 0.924                         & 0.929                         & 0.935                         & 0.940                         & 0.944                         & 0.011 \\
Text Only             & 0.938 & 0.941 & 0.942 & 0.942 & 0.953 & 0.001 \\
Pocket Only           & 0.833                         & 0.838                         & 0.849                         & 0.858                         & 0.860                         & 0.020 \\
Pocket+Text           & 0.929 & 0.932 & 0.937 & 0.940 & 0.948 & 0.008 \\
SG only               & 0.821                         & 0.837                         & 0.844                         & 0.849                         & 0.856                         & 0.012 \\
SG+Text               & 0.957 & 0.961 & 0.963 & 0.965 & 0.970 & 0.004 \\
SG+Pocket             & 0.860                         & 0.866                         & 0.875                         & 0.879                         & 0.882                         & 0.013 \\
OneProt-4             & 0.949 & 0.952 & 0.954 & 0.956 & 0.956 & 0.004 \\
ST only               & 0.882                         & 0.885                         & 0.889                         & 0.892                         & 0.902                         & 0.007 \\
ST+Text               & 0.916 & 0.920 & 0.925 & 0.934 & 0.943 & 0.014 \\
ST+Pocket             & 0.824                         & 0.829                         & 0.841                         & 0.846                         & 0.849                         & 0.017 \\
ST+Pocket+Text        & 0.930 & 0.938 & 0.939 & 0.940 & 0.949 & 0.002 \\
ST+SG                 & 0.889                         & 0.896                         & 0.902                         & 0.927                         & 0.930                         & 0.031 \\
ST+SG+Text            & 0.920 & 0.928 & 0.932 & 0.933 & 0.942 & 0.004 \\
ST+SG+Pocket          & 0.834                         & 0.841                         & 0.862                         & 0.867                         & 0.873                         & 0.026 \\
ProTrek-35M           & 0.935 & 0.940 & 0.944 & 0.950 & 0.954 & 0.011 \\
ProTrek-650M          & 0.962                         & 0.968                         & 0.972                         & 0.973                         & 0.977                         & 0.005 \\
ESM-2                 & 0.934 & 0.938 & 0.941 & 0.947 & 0.952 & 0.010 \\
SaProt                & 0.863                         & 0.936                         & 0.939                         & 0.942                         & 0.944                         & 0.005 \\
ESM-3                 & 0.912 & 0.914 & 0.920 & 0.924 & 0.943 & 0.009 \\
ESM-IF                & 0.854                         & 0.856                         & 0.860                         & 0.863                         & 0.871                         & 0.007 \\
OpenFold              & 0.910 & 0.912 & 0.925 & 0.929 & 0.933 & 0.017 \\
\hline
\textbf{Metal Ion Binding AUC} &                               &                               &                               &                               &                               &       \\
\textbf{AUC }                     & \textbf{Min }                          & \textbf{Q1}                            & \textbf{Median}                        & \textbf{Q3}                            & \textbf{Max}                           & \textbf{IQR}   \\
\hline
OnepProt-5            & 0.919                         & 0.926                         & 0.930                         & 0.934                         & 0.940                         & 0.008 \\
Text Only             & 0.938                         & 0.941                         & 0.942                         & 0.951                         & 0.956                         & 0.010 \\
Pocket Only           & 0.825                         & 0.838                         & 0.845                         & 0.852                         & 0.859                         & 0.014 \\
Pocket+Text           & 0.929                         & 0.932                         & 0.937                         & 0.940                         & 0.948                         & 0.008 \\
SG only               & 0.809                         & 0.832                         & 0.846                         & 0.855                         & 0.858                         & 0.023 \\
SG+Text               & 0.951                         & 0.961                         & 0.963                         & 0.965                         & 0.970                         & 0.004 \\
SG+Pocket             & 0.860                         & 0.866                         & 0.875                         & 0.879                         & 0.882                         & 0.013 \\
OneProt-4             & 0.938                         & 0.940                         & 0.942                         & 0.944                         & 0.952                         & 0.005 \\
ST only               & 0.874                         & 0.882                         & 0.885                         & 0.892                         & 0.902                         & 0.010 \\
ST+Text               & 0.916                         & 0.920                         & 0.922                         & 0.928                         & 0.965                         & 0.007 \\
ST+Pocket             & 0.813                         & 0.826                         & 0.844                         & 0.865                         & 0.876                         & 0.040 \\
ST+Pocket+Text        & 0.933                         & 0.938                         & 0.939                         & 0.943                         & 0.949                         & 0.005 \\
ST+SG                 & 0.889                         & 0.895                         & 0.913                         & 0.925                         & 0.930                         & 0.030 \\
ST+SG+Text            & 0.928                         & 0.933                         & 0.934                         & 0.940                         & 0.945                         & 0.007 \\
ST+SG+Pocket          & 0.836                         & 0.859                         & 0.867                         & 0.872                         & 0.881                         & 0.012 \\
ProTrek-35M           & 0.926                         & 0.936                         & 0.938                         & 0.945                         & 0.954                         & 0.009 \\
ProTrek-650M          & 0.963                         & 0.970                         & 0.973                         & 0.974                         & 0.977                         & 0.003 \\
ESM-2                 & 0.937                         & 0.939                         & 0.947                         & 0.948                         & 0.952                         & 0.009 \\
SaProt                & 0.863                         & 0.936                         & 0.939                         & 0.940                         & 0.941                         & 0.004 \\
ESM-3                 & 0.901                         & 0.914                         & 0.920                         & 0.925                         & 0.943                         & 0.011 \\
ESM-IF                & 0.844                         & 0.852                         & 0.855                         & 0.858                         & 0.866                         & 0.006 \\
OpenFold              & 0.914                         & 0.919                         & 0.926                         & 0.929                         & 0.933                         & 0.010\\
\hline
\end{tabular}
}
\end{table}
%\subsection{ProSPECCTs}\label{supp:prosspects}
}

\end{appendix}

%\clearpage
%\bibliographystyle{plain}
%\bibliography{reference_supp}

%\end{appendices}

\end{document}